\pdfoutput=1
\documentclass[a4paper,fleqn]{cas-dc}
\ExplSyntaxOn
\cs_gset:Npn \__first_footerline:
  { \group_begin: \small \sffamily \__short_authors: \group_end: }
\ExplSyntaxOff

\usepackage[square, numbers]{natbib}
\usepackage{tabularx}

\def\tsc#1{\csdef{#1}{\textsc{\lowercase{#1}}\xspace}}
\tsc{WGM}
\tsc{QE}
\tsc{EP}
\tsc{PMS}
\tsc{BEC}
\tsc{DE}

\begin{document}
\let\WriteBookmarks\relax
\def\floatpagepagefraction{1}
\def\textpagefraction{.001}

\shorttitle{SolarGAN}

\shortauthors{Yufei Zhang et~al.}

\title [mode = title]{SolarGAN: Synthetic Annual Solar Irradiance Time Series on Urban Building Facades via Deep Generative Networks}                      


%
\author[1]{Yufei Zhang}






    
\address[1]{{Chair of Architecture and Building Systems (A/S), ETH Zurich },{Stefano-Franscini-Platz 1},{8093 Zurich },{Switzerland}}

\author[1]{Arno Schlueter}

\author[1]{Christoph Waibel}





\begin{abstract}
Building Integrated Photovoltaics (BIPV) is a promising technology to decarbonize urban energy systems via harnessing solar energy available on building envelopes. While methods to assess solar irradiation, especially on rooftops, are well established, the assessment on building facades usually involves a higher effort due to more complex urban features and obstructions. The drawback of existing physics-based simulation programs are that they require significant manual modelling effort and computing time for generating time resolved deterministic results. Yet, solar irradiation is highly intermittent and representing its inherent uncertainty may be required for designing robust BIPV energy systems. Targeting on these drawbacks, this paper proposes a data-driven model based on Deep Generative Networks (DGN) to efficiently generate high-fidelity stochastic ensembles of annual hourly solar irradiance time series on building facades with uncompromised spatiotemporal resolution at the urban scale. The only input required are easily obtainable, simple fisheye images as categorical shading masks captured from 3D models. In principle, even actual photographs of urban contexts can be utilized, given they are semantically segmented. Our validations exemplify the high fidelity of the generated time series when compared to the physics-based simulator. To demonstrate the model's relevance for urban energy planning, we showcase its potential for generative design by parametrically altering characteristic features of the urban environment and producing corresponding time series on building facades under different climatic contexts in real-time. 
\end{abstract}


\begin{highlights}
\item A generative model using simple fisheye images as inputs to predict solar irradiance time series on building facades in urban environments

\item Fisheye images as categorical shading masks are captured from opensource LOD 1 urban geometry, but in principle may also stem from real photographs

\item Model architecture allows for generation of stochastic ensembles of annual hourly time series, as well as good consistency to the deterministic data

\item Developed a network architecture combining Variational Autoencoder (VAE) and Generative Adversarial Networks (GAN) 

\item Developed model addresses inherent uncertainty of intermittent solar energy and allows parametrically altering urban environmental features 

\end{highlights}


\maketitle

\section{Introduction}
\label{sec:Intro}
\subsection{Urban Solar Modelling: Challenges and Opportunities}
\label{sec:background}

The utilization of solar energy via Photovoltaic (PV), including Building Integrated Photovoltaic (BIPV), is considered a major technology in decarbonizing urban energy systems. Nevertheless, modern cities are formed with complicated mutual shading and reflections, many building surfaces are not able to provide ideal conditions for solar application, which brings additional uncertainties to the intrinsically fluctuating and intermittent solar energy. Therefore, properly evaluating BIPV potential is key for its successful promotion. 

For modelling PV potentials, the process can be divided into different hierarchical steps and scales, starting with the evaluation of annual solar irradiance time-series on tilted (building) surfaces \cite{izquierdo2008method}. At the building-scale, very accurate results with high spatial and temporal resolution can be obtained through labor-intensive 3D modeling and complex simulations in Computer-Aided Design (CAD) software, e.g. using industry-standard engine Radiance \cite{ward1998rendering} and its interface Daysim \cite{reinhart2006tutorial}. Specifically, we need to find local weather data and model the explicit geometries and surface properties of the building envelopes, to accurately calculate the solar beam and diffuse irradiance from the sky subject the surrounding urban context as well as reflections from surrounding buildings. There are several recent studies aiming to speed up the computational process. E.g., Waibel et al. \cite{waibel2017efficient} propose a fast model that only calculate solar irradiance on typical days and use interpolation to complete a year-long result, which still guarantees satisfactory accuracy. While currently the fastest commercial simulation engine, ClimateStudio \cite{climatestudio}, exploits progressive path tracing and hardware acceleration to speed up Radiance. On the other hand, the urban morphology open-source geographical information system (GIS) database released by more and more European cities in recent years, e.g., Level of Detail (LOD) 1 model that contains building footprint polygon and with height information, i.e., minimally necessary geometric input for 3D solar potential evaluation, has greatly eased the trouble of 3D modeling. 

However, even with these advances, accurate simulations are still not obtainable within real-time, especially at the urban scale. Urban-scale terrain and building geometry information is often operated with GIS platforms. Compared to CAD software with powerful capabilities of 3D operations, GIS platform-based simulation methods are often limited to 2D. To consider the influence of urban context, several recent studies have introduced methods based on the classical Sky View Factor (SVF) or shadow mask \cite{oke2017urban, liang2020solar3d, middel2017sky}. Requiring only low computational cost operations such as projection and screenshot, these methods cleverly obtain 2D circular fisheye binary masks to represent the skydome of a sensor-point and reflect the 3D shadowing effect of the surrounding urban context. However, these methods are still unable to reflect the mutual reflections brought by the urban context, which has a critical impact on the vertical surface. Therefore, at the urban scale, it seems inevitable to sacrifice spatial resolution and accuracy to ensure a reasonable simulation time.

In addition, the results of these simulations are often deterministic, whereas urban solar potential is often subject to uncertainties due to fluctuating solar irradiance and several unavailable urban context factors, e.g, windows, vegetations, etc. Regarding weathers, a common strategy is to synthesize Typical Meteorological Year (TMY) weather files from weather data in the several past years. Regarding urban context factors, we usually parametrize them as Window-Wall-Ratio (WWR), vegetation-coverage, etc, and run multiple simulations under a series of settings of these parameters. Peronato et al. \cite{peronato2018toolkit} further propose a method to improve the design of TMY to synthesize both typical and extreme weather scenarios and conduct multiple simulations. However, as a single simulation is already rather time-consuming, these strategies to cope with uncertainties relying on multiple simulation are usually limited in practice as well.

Another alternative is to compensate the simulation approaches via rapid data-driven surrogate models for large-scale applications.  This approach usually exploits some types of machine-learning models trained on simulated ground-truth. E.g., Assouline et al. \cite{assouline2017quantifying} use a 2-stage Support Vector Machine regression model for weather data spatial interpolation and urban BIPV potential estimation. Nault et al. \cite{nault2017predictive} present another workflow to exploit Gaussian Process models for the passive solar potential prediction using simulated results under both randomly generated building geometry blocks and realistic urban morphology for training. Except for the computational cost, both models allow for prediction with empirical intervals, respectively, which is another advantage over the deterministic simulations. In addition, Walch et al. \cite{walch2020fast} conduct a principled benchmark study on five common machine-learning models for urban solar potential prediction and identify Random Forests could achieve the best accuracy. 

Nevertheless, these surrogate models usually lack spatial-temporal resolution by nature. they usually predict the annual cumulative solar irradiation and have to rely on manually designed features to indicate urban context condition, e.g. street width, average building height, etc. These studies therefore generally focus on rooftop PV potential, which is relatively easy to predict. However, the BIPV potential on building facades, which could be several times larger in area than the roofs in modern cities, still lack focused data-driven studies due to their more complex and heterogeneous spatial distribution and temporal variations. To fill this gap, we need to develop a new approach that simultaneously ensures speed, accuracy, and spatial-temporal resolution at the urban-scale.

Recently, the machine learning and artificial intelligence community has brought many exciting new progresses. Numerous Deep Neural Networks (DNN) can already handle complex and high-dimensional data, such as imagery and time-series data. Especially, Deep Generative Networks (DGN) can even create such kinds of data. In the energy sector, DGNs have already been used to generate synthetic energy renewable energy yield or building energy consumption time-series data after being trained on simulated ground-truth data. \cite{chen2018model, fekri2020generating, zhang2018generative, khayatian2021using, baasch2021conditional, dong2022data}. These energy time-series data generators first overcome the lack of time resolution, and can generate high-fidelity but stochastic samples based on the given conditions, e.g., weather, date, value range, etc. These advantages promise that DGN-based models could become the perfect surrogate for high-resolution but computationally expensive simulation models, and even introduce desired variability into the deterministic ground-truth to account for the intermittent renewable resources. Nevertheless, specific to the urban solar modelling, the urban context condition should be ideally aligned to each sensor-point and represented as imagery data to guarantee sufficient spatial-resolution. To the best of the author's knowledge, conditional time-series generation DGN models that could deal with high-dimensional imagery data as given conditions are not reported in the existing literature. On the other hand, the recent advances of generic DGN models already exhibit appealing potentials that via combining several established DGN models together, we could address this difficulty and develop a high-resolution and high-fidelity surrogate model for stochastic solar irradiance time-series data generation, which is the research focus of this study.

\subsection{Key Contributions}
\label{sec:contri}

In general, we propose a model, SolarGAN, which exploits several established DGNs to generate a stochastic ensemble of realistic annual hourly solar profiles on arbitrary urban locations. It has sensor-point level spatial resolution and hourly-level temporal resolution, both are as high as physics-based simulation engines. It also exhibits potential for rapid parametric studies on some essential influencing factors of the urban context, e.g. Window-Wall-Ratio (WWR). In addition, our model requires only easily accessible data input and account for 3D solar potential evaluation only via a simple fisheye image captured from the open-source LOD1 urban geometry model. Therefore it has great generality and broad application prospect. 

In this paper,  We details how such design of the data pipeline as well as model architecture fundamentally guarantee its high fidelity and high spatial-temporal resolution. We also elaborate interesting research topics that could enable this prototype to be deployed in reality; as well as how it could be extended into an end-to-end solution that deal with the real world data, when combining with other established urban scenario segmentation applications.

The remainder of this paper is organized as follows: in section \ref{sec:basics_dgn} we briefly introduce some basics of relevant DGN models applied in this study as well as why and how should we combine several DGNs. Then, section \ref{sec:methodology} details the methodology of this study, including the workflow of the simulated dataset preparation and data processing, and detailed explanations of the model architecture with theoretic background and mindset presented. In section \ref{sec:results} we present the results and evaluate the model performance in terms of generation fidelity via quantitative statistical evaluation as well as its potential of parametric study via qualitative samples demonstration. In section \ref{sec:discussion}, we further elaborate how this model can be further generalized to deal with real-world data and transferred to other related tasks in the future studies, as discussion, which is followed by our final conclusions.

\section{Basics of Deep Generative Networks (DGN)}
\label{sec:basics_dgn}

The general target of DGN models is to approximate complex, ultra-high dimensional data distributions, e.g., imagery and (as in our case) time series data. Via sampling from the approximated distribution, we could simply generate entirely new samples from it that are realistic but do not exist in the training dataset. Thanks to this fitting capability, DGN models could even achieve other related goals, e.g. comprehending, compressing, and even modifying the original imagery and time-series data. Variational Autoencoder (VAE) \cite{kingma2013auto} and Generative Adversarial Networks (GAN) \cite{goodfellow2014generative} are the most popular DGNs at the moment, both developed originally for computer vision tasks but also extended to time-series generation tasks. GAN is specialized in generating high-fidelity samples, and therefore becomes backbone architecture in most of the conditional time-series generator. VAE is generally behind in generation quality, but very good at compressing high-dimensional data and extracting representations from them. When combined \cite{chen2016infogan, donahue2016adversarial, isola2017image,larsen2016autoencoding,zhu2017unpaired,lee2020high}, those DGN models can perform a variety of incredible tasks such as image editing and data modality translation, such as translate urban context imagery data to solar irradiance time-series in our case,  etc. We will introduce the basic ideas of VAE and GAN and their important variants as well as why and how should we combine them, respectively.

\subsection{Variational Autoencoder (VAE)}
\label{sec:vae}

As the name indicates, VAE has interesting connection to a classical DNN architecture, Autoencoder (AE) \cite{hinton2006reducing}, which consists of two DNNs, called encoder and decoder, connected by a layer of latent space. AE is originally developed to extract compact information from high-dimensional and complicated data, e.g., images, in a totally unsupervised manner, i.e., unsupervised dimension reduction. Specifically, given an image, we first use the encoder to compress it to a low-dimensional representation vector in the latent space and then feed it to the decoder to reconstruct a sample that closely resembles the input image. The only training goal of the canonical AE is to minimize this reconstruction error. However, in this case, the latent space might not have many desired properties, such as continuity, disentanglement, etc., which makes it hard to randomly draw vectors from the latent space and feed it into the decoder to generate realistic new samples.

Although the theoretical basis of the VAE model is Variational Bayesian Inference \cite{kingma2013auto}, which is not directly related to AE, its proposed model architecture is still very similar, and its loss function could be empirically understood as an additional regularization term that is added to the reconstruction error term, as a constraint to bring the latent space closer to the standard multivariate normal distribution and therefore obtain those aforementioned properties desired and enables its capability of new sample generation, as shown in Eq. \ref{eq_beta_vae}. In its variant $\beta$-VAE \cite{higgins2016beta}, this regularization term is further strengthened by a factor $\beta$, making each dimension of the latent space more disentangled. Therefore, correspond to critical and interpretable aspects of a sample. E.g., for generic images, some latent dimension might correspond to color, brightness, perspective, etc. Clearly, Using VAE-based models to extract low-dimensional latent representations of complex urban context imagery data could be a effective strategy to turn urban context condition into the format that most of the conditional time-series generators could handle.

\begin{equation}
    \begin{aligned}
    \max _{\theta,\phi}\mathcal{L}_{\beta-\mathrm{VAE}} = &\underbrace{\mathbb{E}_{\mathbf{z} \sim q_{\phi}(\mathbf{z} \mid \mathbf{x})}[\log p_{\theta}(\mathbf{x} \mid \mathbf{z})]}_{\text {reconstruction error }}\\
    &-\underbrace{\beta D_{\mathrm{KL}}\left(q_{\phi}(\mathbf{z} \mid \mathbf{x}) \| p(\mathbf{z})\right)}_{\text {regularization }}\label{eq_beta_vae}
    \end{aligned}
\end{equation}

\subsection{Generative Adverserial Networks (GAN)}
\label{sec:gan}

GAN \cite{goodfellow2014generative} also consists of two DNNs, namely a generator and a discriminator. Similar to the decoder in VAE models, the generator $G$ also transforms a random vector from a latent space that obeys a standard Gaussian distribution into a realistic generative sample to fool the discriminator. While the discriminator $D$ also tries to discriminate the generative samples. $G$ and $D$ act as each other's adversaries and they are trained alternatively to form a $\min\max$ game. At the time when this game finally reaches Nash equilibrium, one can expect that this generator, which can almost fool a very powerful discriminator, is already capable to generate very realistic samples. In practice, what we usually use is actually its variant, conditional GAN (cGAN) \cite{mirza2014conditional}, as shown in Eq. \ref{eq_cgan}, where an additional vector of given condition $c$ is fed into $G$ together with the random latent vector, allowing us to control the generated samples. Similarly, $D$ will also consider the generated samples' fidelity and consistency to the given condition when doing discrimination. As many existing studies \cite{chen2018model, fekri2020generating, zhang2018generative, khayatian2021using, baasch2021conditional, dong2022data}, cGAN becomes the downstream backbone architecture of our conditional solar irradiance time-series generation task.

\begin{equation}
    \min _{G} \max _{D} \mathbb{E}_{x \sim P_{r}}[\log D(c,x)]+\mathbb{E}_{z \sim p_(z)}[\log (1-D(c,G(z)))]\label{eq_cgan}
\end{equation}

GAN models are known for its extraordinary generation quality. One of the reason is  that when fitting the complex and high-dimensional target distribution, GAN models do not require strong prior assumptions on the probability distribution model of the data, which is, however, inevitable in VAE models and always lead to compromised generation performance. More specifically, the loss function ensures that the canonical GAN minimizes the Jensen-Shannon Divergence (JSD) between the probability distribution of the generated and the ground-truth data without knowing their explicit probability distribution models. In some subsequent variants of the GAN, this metric is replaced with the Wasserstein distance with corresponding gradient penalty term added to the loss function  \cite{gulrajani2017improved} (WGAN-GP) that makes training more stable, as shown in Eq. \ref{eq_wgan_g}, \ref{eq_wgan_d}.  Our backbone model architecture for solar irradiance generation is also based on the conditional time-series WGAN-GP model proposed by \cite{lin2020using} called DoppelGANger, which will be detailed in Section \ref{sec:DGNs}.

\begin{equation}
    \begin{aligned}
    \max _{D} &\underbrace{\mathbb{E}_{x \sim P_{r}}[D(x)]-\mathbb{E}_{\tilde{x} \sim P_{g}}[D(\tilde{x})]}_{\text {Wasserstein discriminator objective }}+\\
    &\lambda \underbrace{\mathbb{E}_{\hat{x} \sim P_{r} \, or \, P_{g}}\left[\left(\left\|\nabla_{\hat{x}} D(\hat{x})\right\|_{2}-1\right)^{2}\right]}_{\text {Gradient Penalty}}\label{eq_wgan_d}
    \end{aligned}
\end{equation}

\begin{equation}
    \min _{G} -\mathbb{E}_{\tilde{x} \sim P_{g}}[\log (D(\tilde{x}))]\label{eq_wgan_g}
\end{equation}

\subsection{Combining VAE and GAN for image editing and modality translation}
\label{sec:combine}

In addition to generating realistic samples, using GAN for image-editing tasks is also a very active research field. In fact, the ability of image-editing is critical for developing DGN-based urban solar prediction surrogate model models, which is because the urban context information represented by imagery data is often accompanied by many uncertainties. Especially when this imagery data is captured from low LOD, architectural components such as windows and shading elements as well as urban vegetation are usually simplified or overlooked, and the spatial location of sensor-points may also have some deviations. If our model is able to adjust these details when being trained with an imagery dataset containing different LODs and different viewpoint locations, then we could parametrically study these uncertainties as in 3D CAD software. We believe that this capability of parametric study is essential for a transversal, scalable, and flexible urban solar prediction surrogate model. 

Interestingly, many studies in this field combine the conditional generator with an encoder, or an entire VAE model. e.g. \cite{chen2016infogan, donahue2016adversarial, isola2017image,larsen2016autoencoding,zhu2017unpaired,lee2020high}, As the high-dimensional base input image could be encoded to a low-dimensional vector of latent representations as the generation conditions. Desired modifications will also be imposed by means of modifying the dimensions that corresponds to the aspects to be edited. InfoGAN proposed by Chen et al. \cite{chen2016infogan} is a representative work. An additional encoder is introduced to the canonical GAN model with an additional regularization term added to the canonical loss function. The main idea is to maximize the mutual information between a generation sample and its encoded latent vector. Therefore, the latent vector could be used as conditional input to control specific aspects of the generated samples. Lee et al. \cite{lee2020high} proposed an variant of InfoGAN, called Information Disstilatioin GAN (ID-GAN), which use a well-trained encoder from a $\beta$-VAE instead of training an encoder from the scratch, to capitalize on the neat disentangled latent representations from $\beta$-VAE and ensure better generation quality. This model architecture allows us to principledly address the task of extracting low-dimensional image representation and image-editing at the same time, and therefore becomes part of our upstream backbone model architecture to deal with the imagery input representing the urban context information.

When combine the upstream model to deal with imagery data and the downstream model to generate time-series data conditionally, our model is able to conduct a mapping from urban context information in imagery data, together with other relevant conditions, to solar irradiance time-series data. To the best of our knowledge, this multimodal architecture has no precedent in data-driven urban solar study, but its design is coherent to the basic concept of crossmodal translation \cite{zhang2020multimodal}. We also noticed the work of Paletta et al. \cite{paletta2021benchmarking, paletta2020convolutional} that use fisheye photos obtained from the sky imager to predict ultra-short-term solar irradiance exhibit similar mindset and data pipeline, although they did not apply DGNs. The detailed explanation of DGNs applied in our proposed model could be found in Section \ref{sec:DGNs} in the section of methodology.

\section{Methodology}
\label{sec:methodology}

\begin{figure*}[h]
	\begin{center}
	\includegraphics[width=1\textwidth]{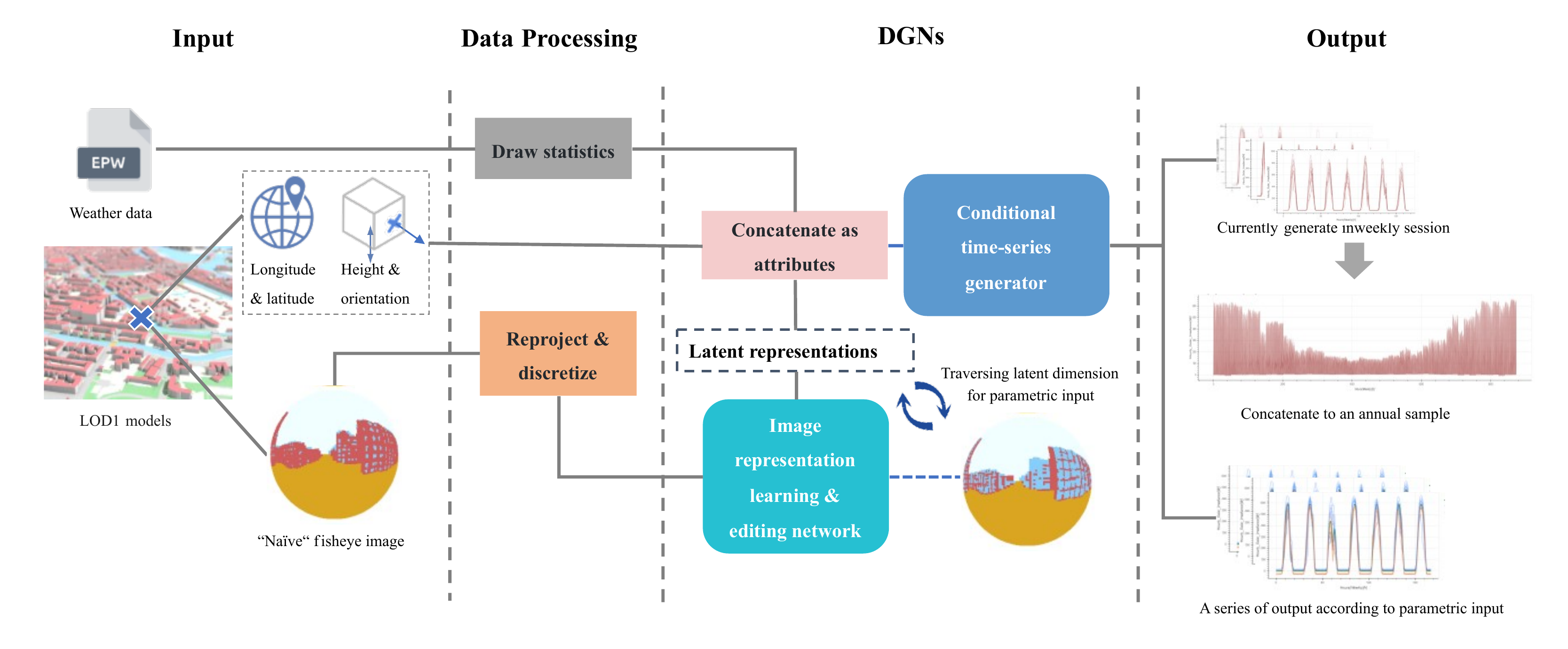}
    \caption{\label{fig:overview}Overview of the data pipeline and the model architecture}
  	\end{center}
\end{figure*}

To develop such a high fidelity and high resolution surrogate model, we propose a workflow that contains 3 work-packages spanning from data acquisition, processing, to the model design based on several state-of-the-art DGNs, as shown in Fig. \ref{fig:overview}.

\subsection{Dataset preparation}
\label{sec:dataset}

\begin{figure*}[h]
	\begin{center}
	\includegraphics[width=1\textwidth]{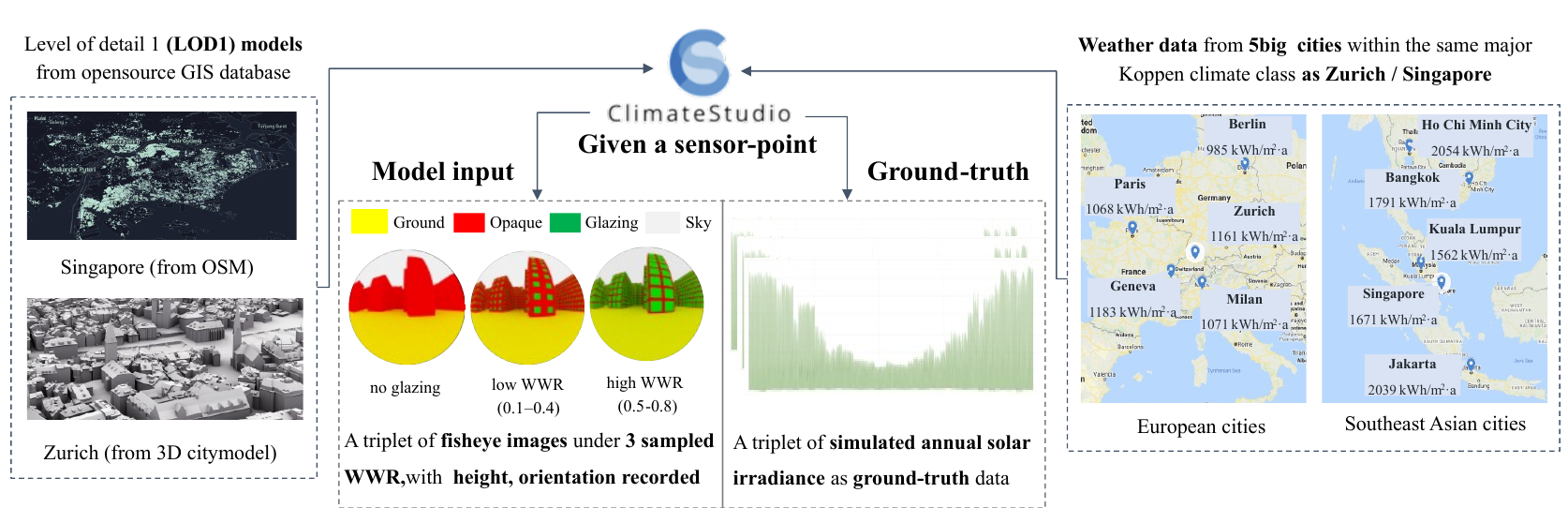}
    \caption{\label{fig:dataset}Dataset preparation workflow}
  	\end{center}
\end{figure*}

We prepare the dataset using urban LOD1 geometries as well as weather files from both European and Southeast Asian cities, to fundamentally guarantee the model's generality. Zurich and Singapore become representative cities that provide LOD1 models with five other cities located in the same major Köppen climate region respectively, as shown in Fig. \ref{fig:dataset}. For the city Zurich, the Civil engineering and disposal department of the city Zurich has open sourced their LOD1 3D city model from Geomatik + Vermessung (GeoZ) \cite{3Dstadt}. While for the city Singapore, although there are also a lot of initiatives of detailed urban modelling led by the local municipality,  currently the most easily accessible data source is still from the world-wide open-source urban information database, OpenStreetMap (OSM) \cite{OSM}.

These LOD1 models in GIS format can be imported as 3D models with the help of Rhino \& Grasshopper plugin ``@it'' \cite{@it}. As already explained, the simulation of the solar irradiance is specific to a sensor-point on the building surface. Therefore, we sampled 5'000 sensor-points on the facades from the Zurich and Singapore LOD1 models, respectively. There are two reasons for choosing only the sensor-points on the façade: first, we find that the majority of the sensor-points on the rooftop are mostly unobstructed, resulting in less valuable samples for training. Second, many methods for rooftop solar irradiance prediction already exist, while vertical surfaces / facades are less well studied in this context. Nevertheless, it is important to note that our proposed approach is still feasible to both rooftop and facade points.

We simulate the ground-truth solar irradiance time-series data for these sensor-points via the simulation engine ClimateStudio \cite{climatestudio}, it also has powerful rendering features that allows us to capture imagery data representing the individual urban context condition of these sensor-points. It is necessary to first explain how we decides the input format of these imagery data.

In general, what we captured for each sensor-point is so-called multi-categorical shadow masks, which are basically circular fisheye images as shown in Fig.  \ref{fig:dataset}. It is modified from the classical shadow masks that always point to the zenith and only have binary values. 

To get such a multi-categorical shadow mask, we need to create a sunlit hemisphere along the normal vector of the surface where the test-point is located (instead of still creating a skydome along the zenith) and then perform a hemispheric fisheye projection. For different categories of surrounding elements, we could apply different standard colors via assign them to different layers, and obtain minimum quality rendering of fisheye images, which takes only a few seconds for a test-point.

We also generate a triplet of ground-truth solar irradiance as well as fisheye images for each sensor-point with three WWR levels for the surrounding building surfaces during simulation, as shown in Fig.  \ref{fig:dataset}. As explained in Section \ref{sec:combine}, this intentional design of the dataset is a preliminary attempt to introduce slightly different level of LODs, which  enables us to evaluate our model's capability of parametric study on the influence of windows. Table \ref{tab:simu_data_overview} gives an overview of the entire simulated dataset.

\begin{table*}
\caption{Simulation datasets overview}
\label{tab:simu_data_overview}

\begin{tabular}{cccccc}
\hline
City
  geometry & \#sensor-points         & \begin{tabular}[c]{@{}c@{}}\#image\\~samples\end{tabular} & \begin{tabular}[c]{@{}c@{}}image\\~size\end{tabular} & \begin{tabular}[c]{@{}c@{}}\#ground-truth\\~samples\end{tabular} & \#train /
  \#test    \\ 
\hline
Zurich          & \multirow{2}{*}{5000} & \multirow{2}{*}{15000}                                    & \multirow{2}{*}{512}                                 & \multirow{2}{*}{75000}                                           & \multirow{2}{*}{4:1}  \\
Singapore       &                       &                                                           &                                                      &                                                                  &                       \\
\hline
\end{tabular}
\end{table*}

\subsection{Data processing}
\label{sec:processing}

 \begin{figure*}[h]
	\begin{center}
	\includegraphics[width=1\textwidth]{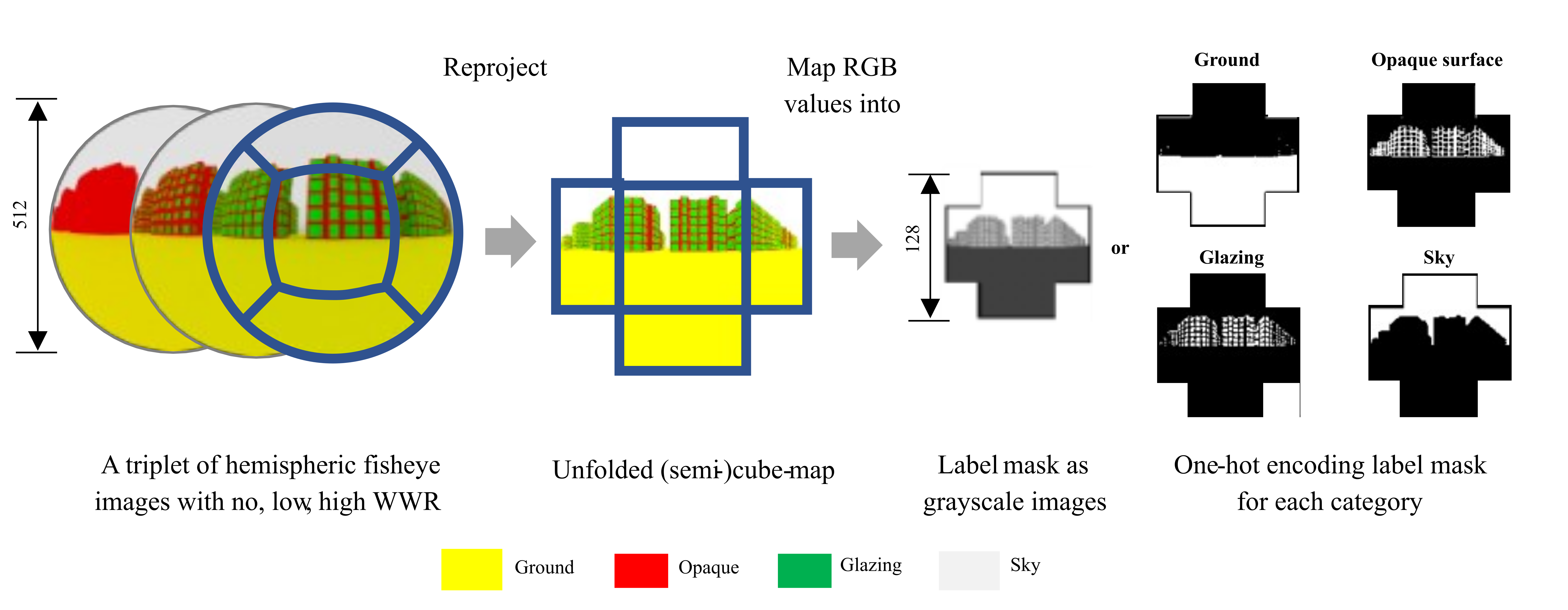}
    \caption{\label{fig:im_pocess} Illustration of the image processing}
  	\end{center}
\end{figure*}

As shown in Fig. \ref{fig:im_pocess}, these fisheye categorical mask images are further reprojected to unfolded semi-cube-maps, with one-hot-encoded (multiple binarized channels) or order-encoded (single-channel grayscale image) pixel-wise categorical information. Such image formats are concise while the image features remain identifiable after shifting and scaling, which is preferred by Convolutional Neural Networks (CNN) that are specialized to deal with images. 

As per the weather data input, i.e., Direct Normal Irradiance (DNI) and Diffuse Horizontal Irradiance (DHI), instead of feeding them in as time-series, we note that only drawing some statistics, e.g., peak and average values, could be a simple but effective way that avoids model complexity \cite{baasch2021conditional} and allows for more flexible generation that do not strictly follow the deterministic weather patterns in the weather file. As a time-series with 8760 steps is too long for any time-series DGN, we experimentally let the model generate patches by weekly sliding windows. Therefore, the weather statistics are also drawn in weekly segments. As shown in Fig. \ref{fig:overview}, except for images and weather statistics, auxiliary conditional input of a sensor-point for generation also includes longitude and latitude of the weather file location, as well as height and surface normal vector of the sensor-point. Both are easily acquirable from the LOD1 model.

Regarding the weekly patches of solar irradiance to be generated, we further noticed that all the European and Southeast Asian cities have sunlit hours only between 4:00 and 21:00. Therefore, it is sufficient to keep only these 17 time-steps in each day, instead of keeping all 24 time-steps. Therefore, the length of each weekly sample can be reduced to $17 \times 7 = 119$.

Table \ref{tab:model_input_data_overview} gives an overview, which holds for both the Singapore and the Zurich dataset. These attributes and ground-truth and the corresponding cube-map images will be directly used as input to the DGNs.

\begin{table}[width=.8\linewidth,cols=3,pos=h]
\caption{Model input dataset overview (applies to both the Singapore and the Zurich dataset)}
\label{tab:model_input_data_overview}
\begin{tabular}{l|l|l}
\multicolumn{3}{c}{Attributes}                                                                                                                                                                                                                                                 \\ 
\hline
Term                                                                 & Description                                                                                                                                                                             & Dim.    \\ 
\hline
\begin{tabular}[c]{@{}l@{}}\\Longitude \\and latitude\\\end{tabular} & \begin{tabular}[c]{@{}l@{}}\\Location in the weather file\\\end{tabular}                                                                                           & 2             \\
\begin{tabular}[c]{@{}l@{}}\\Height\\\end{tabular}                   & \begin{tabular}[c]{@{}l@{}}\\z-coordinate of the test-point\\\end{tabular}                                                                                                              & 1             \\
\begin{tabular}[c]{@{}l@{}}\\Surface \\ normal\\\end{tabular}           & \begin{tabular}[c]{@{}l@{}}\\x, y component of the surface normal \\ vector given the facade sensor-point \\ (z-component unnecessary)\\\end{tabular}                                   & 2             \\
\begin{tabular}[c]{@{}l@{}}\\Monthly \\ index\\\end{tabular}            & \begin{tabular}[c]{@{}l@{}}\\monthly index of the weekly patch\\\end{tabular}                                                                                                           & 1             \\
\begin{tabular}[c]{@{}l@{}}\\Solar \\ declination\\\end{tabular}        & \begin{tabular}[c]{@{}l@{}}\\calculate using the first day \\ in that month \\\end{tabular}                                                                & 1             \\
\begin{tabular}[c]{@{}l@{}}\\Weather \\ statistics\\\end{tabular}       & \begin{tabular}[c]{@{}l@{}}\\For DNI and DHI of each week:\\ · hourly peak and hourly average \\ · hourly average of the max./min. day \\\end{tabular} & 8             \\ 
\hline
\multicolumn{1}{l}{total}                                            &                                                                                                                                                                                         & 15            \\ 
\hline
\multicolumn{1}{l}{}                                                 & \multicolumn{1}{l}{}                                                                                                                                                                    &               \\
\multicolumn{3}{c}{Ground-truth
  time-series}                                                                                                                                                                                                                                 \\ 
\hline
Term                                                                 & \multicolumn{2}{l}{Description}                                                                                                                                                                         \\ 
\hline
\begin{tabular}[c]{@{}l@{}}\\Sample \\ length\\\end{tabular}            & \multicolumn{2}{l}{\begin{tabular}[c]{@{}l@{}}\\17 time-steps × 7 days = 119\\\end{tabular}}                                                                                                            \\
\begin{tabular}[c]{@{}l@{}}\\Daily \\ duration\\\end{tabular}           & \multicolumn{2}{l}{\begin{tabular}[c]{@{}l@{}}\\4:00 - 21:00, 17 time-steps\\\end{tabular}}                                                                                                             \\
\begin{tabular}[c]{@{}l@{}}\\Dataset \\ size\\\end{tabular}             & \multicolumn{2}{l}{\begin{tabular}[c]{@{}l@{}}\\train: 3120 K weekly samples\\ test: 780 K weekly samples\\\end{tabular}}  \\
\hline
\end{tabular}
\end{table}

\subsection{Designing and training of DGNs}
\label{sec:DGNs}

As shown in Fig. \ref{fig:overview}, our proposed model architecture consists of two main components, i.e., a DGN to deal with imagery urban context information and a DGN to generate solar irradiance conditionally. We would like to elaborate on their detailed design in Section \ref{sec:DNG4images} and \ref{sec:DNG4timeseries}. Then, detailed data pipeline as well as strategies to combines these DGNs will be detailed in Section \ref{sec:DGNtrain}.  

\subsubsection{DGN for images}
\label{sec:DNG4images}

The DGN model architecture to handle images is based on the Information Distillation GAN (ID-GAN), proposed by Lee et al. \cite{lee2020high}. ID-GAN integrates $\beta$-VAE and InfoGAN, the core idea is to first train a lightweight $\beta$-VAE focusing on learning image representations, and then using the well-trained encoder to train a much larger InfoGAN for high-fidelity generation and image-editing. Now, we could obtain a VAE that is dedicated to efficiently extracting meaningful and disentangled image representations, as well as a DGN that is dedicated to generating images based on the image representations extracted by the upstream $\beta$-VAE, either directly from a ground-truth image, or with some specific dimensions of representations modified for the purpose of image-editing. The subsequently trained GAN could still strictly follow the representations extracted by the upstream VAE, ensuring that the generated and edited images are indeed based on a given ground-truth image. This process is similar to passing down the knowledge from one DGN to another, and is also the reason why ID-GAN gets this name.

\begin{equation}
    \begin{aligned}
    \min _{G} \max _{D} & \mathbb{E}_{x \sim P_{r}}[\log D(x)]+\mathbb{E}_{z \sim p_(z)}[\log (1-D(G(z)))] \\ 
    &-\lambda \mathcal{R}_{\text {ID-GAN }}\label{eq_idgan}
    \end{aligned}
\end{equation}

\begin{equation}
    \mathcal{R}_{\mathrm{ID-GAN}}=\mathbb{E}_{c \sim q_{\phi}(c), \tilde{x} \sim P_{g}(\tilde{x}}\left[\log q_{\phi}(c \mid \tilde{x})\right]+H_{q_{\phi}}(c)\label{eq_idgan_elbo}
\end{equation}

The training loss function of the $\beta$-VAE is exactly identical to Eq. \ref{eq_beta_vae}. Then, for the ID-GAN, similar to InfoGAN that is introduced in Section \ref{sec:combine}, its loss function (Eq. \ref{eq_idgan}) is equivalent to add an regularization term (Eq. \ref{eq_idgan_elbo}) to the canonical GAN loss. The calculation of the regularization term relies on a well-trained encoder $q_{\phi}$ from the previously trained $\beta$-VAE. Here, $q_{\phi}(c)=\frac{1}{N} \sum_{i} q_{\phi}\left(c \mid x_{i}\right)$ is the aggregated posterior distribution of the latent vector after encoding, and $H_{q_{\phi}}(c)$ denotes its information entropy, which is usually calculated with this equation:    $H(X)=E_{x \sim p(x)}[-\log p(x)]=-\sum_{x} p(x) \log \frac{1}{p(x)} $

The detailed architecture of the $\beta$-VAE  and ID-GAN have been shown in Fig. \ref{fig:betavae} and \ref{fig:idgan} in appendix, respectively. Both are based on Convolutional Neural Networks (CNN), which is typically used to deal with imagery data. For image generation and editing, we note that treating cube-maps as simple order encoding grayscale images is already sufficient. Therefore ID-GAN generates images with a size of 128 (W) × 128 (H). It will still be transformed to one-hot encoding label masks when being fed into the $\beta$-VAE encoder. The input dimension for the ID-GAN generator is 36, containing 32 dimensional image representations as well as a random latent vector with a length of 4. Table \ref{tab:training} provides more information about the training settings and hyperparameters.

\subsubsection{DGN for time-series}
\label{sec:DNG4timeseries}

Since the $\beta$-VAE encoder could compress an image into a vector of image representations, all the generation conditions for a weekly time-series sample are in standard tabular-data form and could be fed into a conditional time-series GAN. The total dimension of the condition is 47, including 15-dimensional original attributes as well as 32-dimensional image representations. As explained in Section \ref{sec:processing}, the output time-series samples are in weekly patches with only sunlit hours reserved. Their length is 119.

The architecture of the conditional time-series GAN in our case is based on DoppelGANger, proposed by Lin et al. \cite{lin2020using}. The main highlight of this model is to introduce an auxiliary GAN to generate the maximum and minimum values of a time-series sample according to the given conditions. 

Without this auxiliary GAN, the time-series generator has to generate many samples with a large variety of ranges, which is common for solar irradiance samples under different orientations, seasons, or urban contexts. This challenging task probably leads to a problem called "mode collapse", where the generated samples are always trapped within a certain range. The auxiliary GAN actually functions as a clever method of instance normalization, i.e. the time-series generator only needs to generate time-series samples between zero and one. Such a design could significantly improve the quality of conditional generation. Both time-series GAN  and auxiliary GAN in DoppelGANger use a WGAN-GP (Eq. \ref{eq_wgan_d}, \ref{eq_wgan_g}) as their loss functions. When training together, the overall loss function is a linear combination of both, with a factor of $\alpha$ for the auxiliary GAN loss, as shown in Eq.\ref{eq_doppelganger}. 

The specific model architecture of the conditional time-series generator has been shown in Fig. \ref{fig:doppelganger} in appendix. The time-series generator exploits a classical Recurrent Neural Network (RNN) architecture to deal with sequential data, i.e., Long Short-term Memory (LSTM). While both the time-series classifier as well as the auxiliary GAN use vanilla DNN architecture, i.e. Multi-layer Perceptron (MLP).

\begin{equation}
    \min _{G_{aux.}, G} \max _{D_{aux.}\,D} \mathcal{L}(G,D)+\alpha\mathcal{L}_{\text {aux.}}(G_{aux.},D_{aux.})\label{eq_doppelganger}
\end{equation}

Finally, it is important note that because it is very time-consuming to train the model on the entire huge training dataset due to the weekly division (3120 K), we use only a quarter of the training data, i.e. 1000 test-points and 780 K weekly samples for both the Singapore and Zurich dataset, respectively. Table \ref{tab:training} provides detailed information about the training settings and hyperparameters.

\subsubsection{Put everything together}
\label{sec:DGNtrain}

 \begin{figure*}[htbp]
	\begin{center}
	\includegraphics[width=0.8\textwidth]{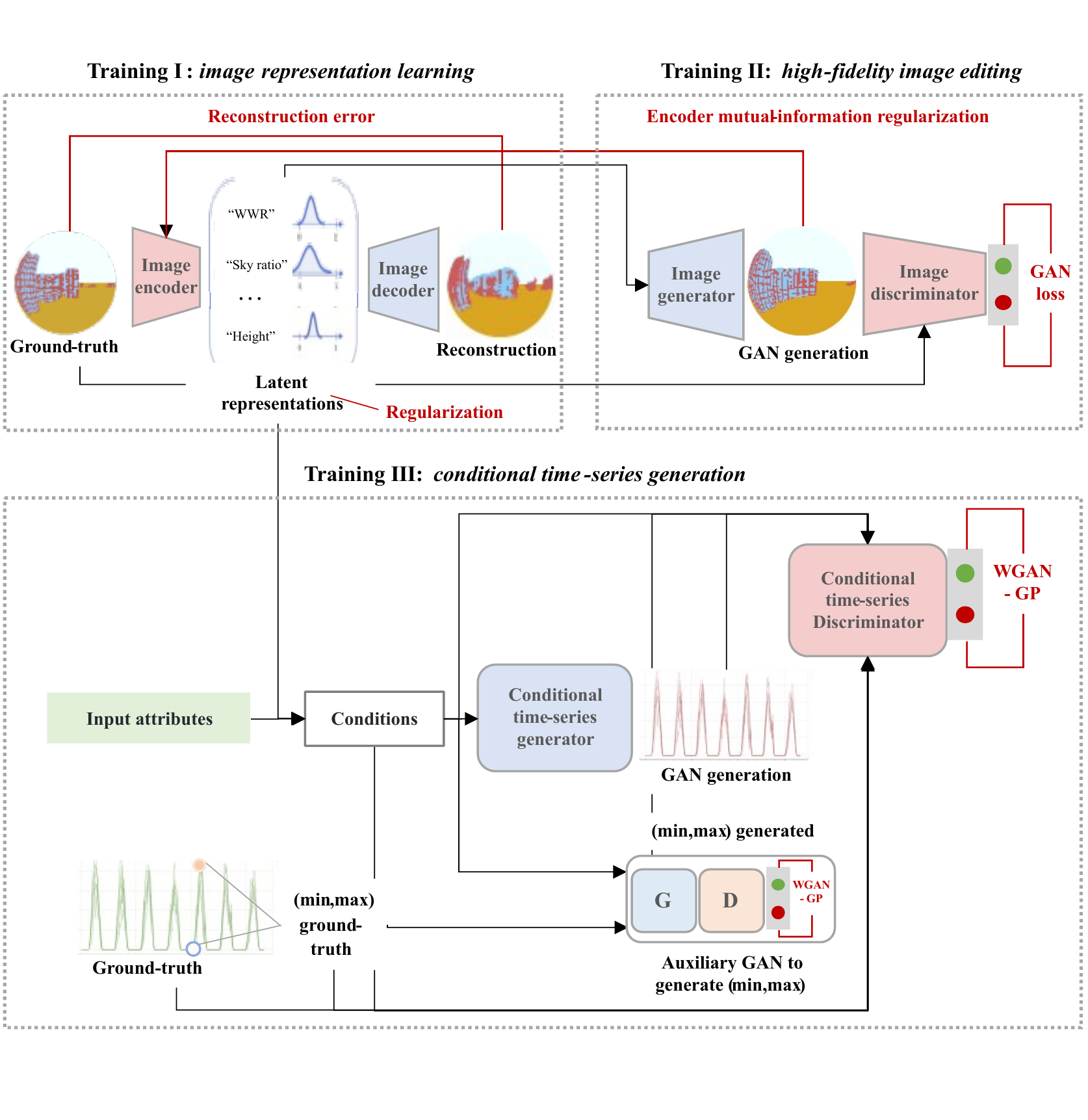}
    \caption{\label{fig:DGNtrain} Illustration of the detailed model architecture and the entire training process in 3 separate stages. Lines and texts in red indicate the loss function terms involved}
  	\end{center}
\end{figure*}

The entire training process and overall architecture of the aforementioned DGNs have been illustrated in Fig. \ref{fig:DGNtrain}. Although these DGNs are trained in three separate stages, the clever design of ID-GAN first helps link image representation learning tightly to image-editing and even enables better respective performances on both tasks than joint training, according to the ablation study by Lee et al. \cite{lee2020high}. In addition, although the representations from the $\beta$-VAE encoder are not optimized along with the training of conditional time-series generation, the cub-map image itself is already in a well-suited form that represents relevant urban context information with respect to solar potential, as it is stemmed from the classical physics-based simulation approaches. Those representations extracted from the urban context images by the encoder, e.g., viewpoint coordinates, object contours, and sky obstruction conditions, are also consistent with the way human experts would like to interpret these images. Therefore, we argue that training separately does make sense in this preliminary research. It could also help us better understand the effects and impacts from individual DGNs. Table \ref{tab:training} summarizes the key facts about the training of these three types of DGNs, including hyperparameters, optimizer settings, training duration, model sizes, etc. For DoppelGANger, it is important to note that we actually train 2 models with the same architecture on both the Singapore and the Zurich dataset, respectively. 

\begin{table*}[htbp]
\renewcommand\arraystretch{1.2}
\small
\centering
\caption{Training overview of DGNs}
\label{tab:training}
\begin{tabular}{c|c|c|c} 
\hline
Term              & $\beta$-VAE                                                    & ID-GAN                                                                                                          & DoppelGANger                                                                                                                                                                                                                          \\ 
\hline
Hyperparameters   & $\beta$ = 3 (Eq.
  \ref{eq_beta_vae})       & $\lambda$ = 0.01 (Eq.
  \ref{eq_idgan})                                    & \begin{tabular}[c]{@{}c@{}}$\alpha$ = 1 (Eq. \ref{eq_doppelganger})\\ $\lambda$ =$\lambda_{aux.}$ = 10 (Eq. \ref{eq_wgan_d})\end{tabular}  \\ 
\hline
Batch size        & 32                                                                              & 64                                                                                                              & 100                                                                                                                                                                                                                                   \\ 
\hline
Optimizer         & Adam \cite{kingma2014adam}
  ($\beta1 = 0.9$) & Rmsprop \cite{hinton2012neural}                                                                & Adam \cite{kingma2014adam}
  ($\beta1 = 0.5$)                                                                                                                                                       \\ 
\hline
Learning rate     & $lr$ = $10^{-4}$                                                               & \begin{tabular}[c]{@{}c@{}}$lr$ = $10^{-4}$\\ (anneal by a factor of 0.9 \\ every $10^{4}$ iterations)\end{tabular} & \begin{tabular}[c]{@{}c@{}}$lr$ = $10^{-3}$\\ ($D$ train 3 steps while\\ ~$G$ train 1 step)\end{tabular}                                                                                                                           \\ 
\hline
Training duration & \begin{tabular}[c]{@{}c@{}}2500 K iterations \\ (68 hours)\end{tabular}         & \begin{tabular}[c]{@{}c@{}}500 K iterations \\ (100 hours)\end{tabular}                                         & \begin{tabular}[c]{@{}c@{}}200 epochs \\(1560 K iterations)\\ (72 hours)\end{tabular}                                                                                                                                                           \\              \hline
Environment       & Pytorch 1.9 \cite{NEURIPS2019_9015}                           & Pytorch 1.9~\cite{NEURIPS2019_9015}                                                          & Tensorflow 1.15 \cite{tensorflow2015-whitepaper}                            
                                                                                                                          \\ 
\hline
Model size~       & 18 MB                                                                           & 238 MB                                                                                                          & 6 MB                                                                                                                                                                                                                                  \\ 
\hline
GPU               & \multicolumn{3}{c}{Nvidia 3090 (memory: 24 GB)}                                                                                                                                                                                                                                                                                                                                                                                           \\
\hline
\end{tabular}
\end{table*}

Lastly, although the training stage involves very complicated model architectures, in the exploitation phase we only need to keep the encoder in $\beta$-VAE and generators in ID-GAN and DoppelGANger. Together with the table \ref{tab:training}, we could observe that the encoder and time-series generator frequently used in the basic workflow are still quite lightweight. Therefore, they are expected to be flexibly deployed on the user side. On the other hand, the much heavier image generator is only involved in the optional workflow for interactions with users. For example, traversing the representation corresponding to the WWR for some images to help users add widows or adjust WWR of the surrounding context. This optional workflow should only affect a few sensor-points and can be considered for remote calls only.

\section{Results}
\label{sec:results}

Next, we use the test dataset to evaluate the performance of our model, SolarGAN, including following aspects:

\begin{enumerate}

\item We first evaluate the model's capability on image representation learning and image-editing. In general, we believe the consistency between the ground-truth samples and reconstructed samples that are up-sampled only from the low-dimensional representations could reflect the quality of the extracted representations. The similar rationale also applies to the evaluation on the image-editing task via comparing the edited images with respect to WWR via traversing the corresponding dimension to the ground-truth images with similar WWR levels. As the images in our case are multi-categorical masks, which is in similar form to another classical computer vision task, semantic segmentation, which aims to do pixel-wise classification for an image. Therefore we use the same metrics, i.e., Pixel Accuracy (PA) and Intersect over Union (IoU), in the semantic segmentation tasks to evaluate the pixel-wise consistency. 

\item We then evaluate the fidelity of the generated solar irradiance time-series samples. We first use several statistical metrics, i.e. Jensen-Shannon Divergence (JSD) and autocorrelaction (AC), to evaluate the consistency between the ground-truth and generated samples with respect to value distribution, short-term and long-term trend. Then, we also examine whether the generated and ground-truth samples are interchangeable to the downstream machine-learning models as an alternative to evaluate the fidelity. Therefore, we trained a time-series classifier to compare its classification performance on the ground-truth and generated samples, using F1 Score as the evaluation metric. 

\item We further qualitatively demonstrate the model's potential of parametric study on several urban context factors, including WWR, which is intentionally introduced to the dataset, as well as viewpoint coordinates, which are prominent variabilities in the dataset that is unsupervisedly discovered by the model. 

\item Finally, we compare the computing time of SolarGAN with the currently fastest solar simulation engine, ClimateStudio, to demonstrate its possible advantages and current limitations in terms of speed.

\end{enumerate}

\subsection{Performance evaluation on images}
\label{sec:eval_images}

\begin{figure*}[htbp]
    \centering
	\includegraphics[width=1\textwidth]{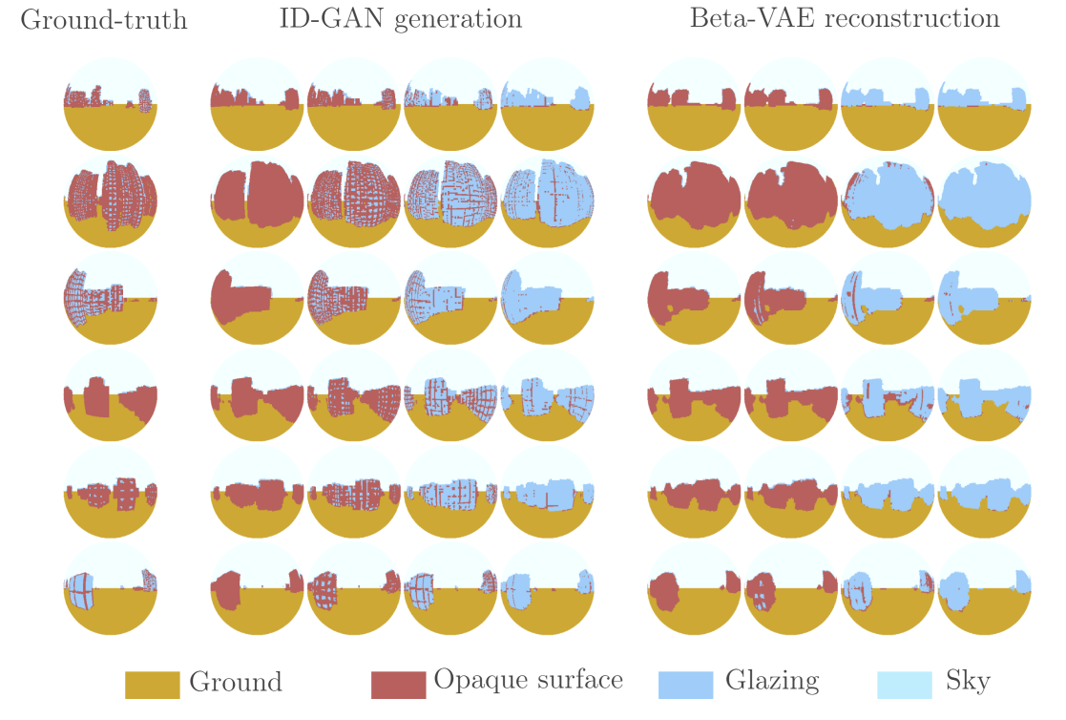}
    \caption{\label{fig:wwr}Traversing the latent representation of WWR. The sample images has been recolored and shown in the fisheye format for more intuitive visual appearance}
\end{figure*}

The quality of image representations reflects how much urban context information the model could extract and compress from the input images, which is crucial to the subsequent conditional time-series generation task.  Therefore, it is definitely necessary to observe the $\beta$-VAE reconstruction performance. In addition, since ID-GAN cannot "see" the ground-truth image directly but only the encoded latent representation, the consistency of its generated samples with the ground-truth images is entirely guided by the encoder of $\beta$-VAE, its generation performance is also relevant and informative. 

As shown in Fig. \ref{fig:wwr}, It could be seen that both $\beta$-VAE and ID-GAN are able to capture global characteristics of the ground-truth images. In general, we could observe the consistency of the generated or reconstructed samples with the ground-truth images. In addition, it could also be observed that the encoder of $\beta$-VAE does effectively associate one dimension of image representations to the actual underlying generative factor, WWR of the surrounding buildings, in a completely unsupervised manner. Although $\beta$-VAE is still unable to generate realistic samples, ID-GAN could generate much more realistic window patterns under a specific latent WWR.

Then, we evaluate the pixel-wise consistency between the generated and ground-truth categorical masks quantitatively, which is equivalent to evaluate a pixel-wise classification task. The simplest evaluation approach is to directly calculate the ratio of correctly classified pixels, called Pixel Accuracy (PA). In binary classification, True positive (TP) and True Negative (TN) are the pixels that are distinguished correctly as target and background, respectively. Whereas False positive (FP) and False Negative (FN) are the rest pixels that are wrongly classified. PA is calculated as Eq. \ref{eq_pa}.

\begin{equation}
   PA=\frac{\# T P+\# T N}{\# T P+\# T N+\# F P+\# F N} \label{eq_pa}
\end{equation}

However, matching pixels exactly according to the positions might be too strict. Intersect over Union (IoU) is a more complex but more informative metric to reflect a model's capability to capture the position and contour of each category of objects, From the perspective of our image generation task, this metric expects that the pixels correctly generated for each category of objects should make up as large a proportion as possible within the union of all pixels of this category from both the generated samples and the ground-truth. Nevertheless, small offset or discrepancy will not significantly degrade this metric. The equation to calculate IoU for one category is given in Eq. \ref{eq_iou}. We could also leverage mean IoU (mIoU) to quantify whether the model has a balanced generative capacity for each category of objects by simply summing and averaging the IoU of each category.

\begin{equation}
   I o U=\frac{\text { Intersection }}{\text { Union }}=\frac{\# T P}{\# T P+\# F P+\# F N} \label{eq_iou}
\end{equation}

\begin{figure}[htbp]
    \centering
	\includegraphics[width=0.45\textwidth]{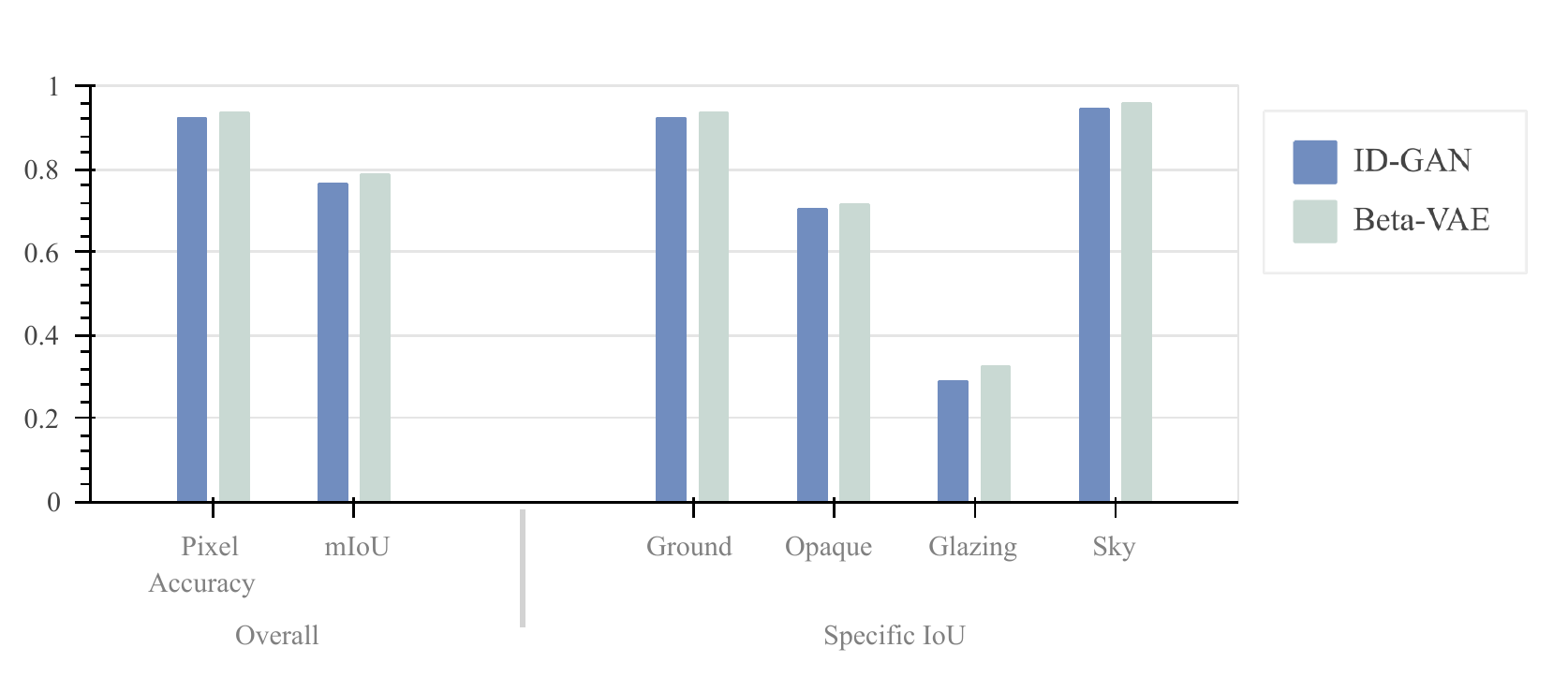}
    \caption{\label{fig:iou}Quantitative evaluation of the $\beta$-VAE reconstruction and ID-GAN generation via PA, mIoU, and IoU of each category}
\end{figure}

These aforementioned metrics all range from zero to one, the larger, the better. We plot the PA, mIoU, as well as IoU of each category for $\beta$-VAE reconstruction and $ID-GAN$ generation together in Fig. \ref{fig:iou}.  In terms of PA, both $\beta$-VAE and $ID-GAN$ exceed 0.9, which could convince the overall consistency. However, the IoU metrics of both models are not that satisfactory, mainly because the IoU of glazing for both models are surprisingly lower to 0.3. This under-performance also seems to affect the IoU of the opaque surfaces, since glazing and opaque surfaces appear complementarily. Moreover, although ID-GAN generates more realistic samples, it does not gain an advantage in terms of consistency. In fact, it can be seen that $\beta$-VAE, which can directly "see" the ground-truth, always has a small lead over ID-GAN in all metrics, although the gaps between their performance metrics are not large.

\begin{figure}[htbp]
    \centering
	\includegraphics[width=0.45\textwidth]{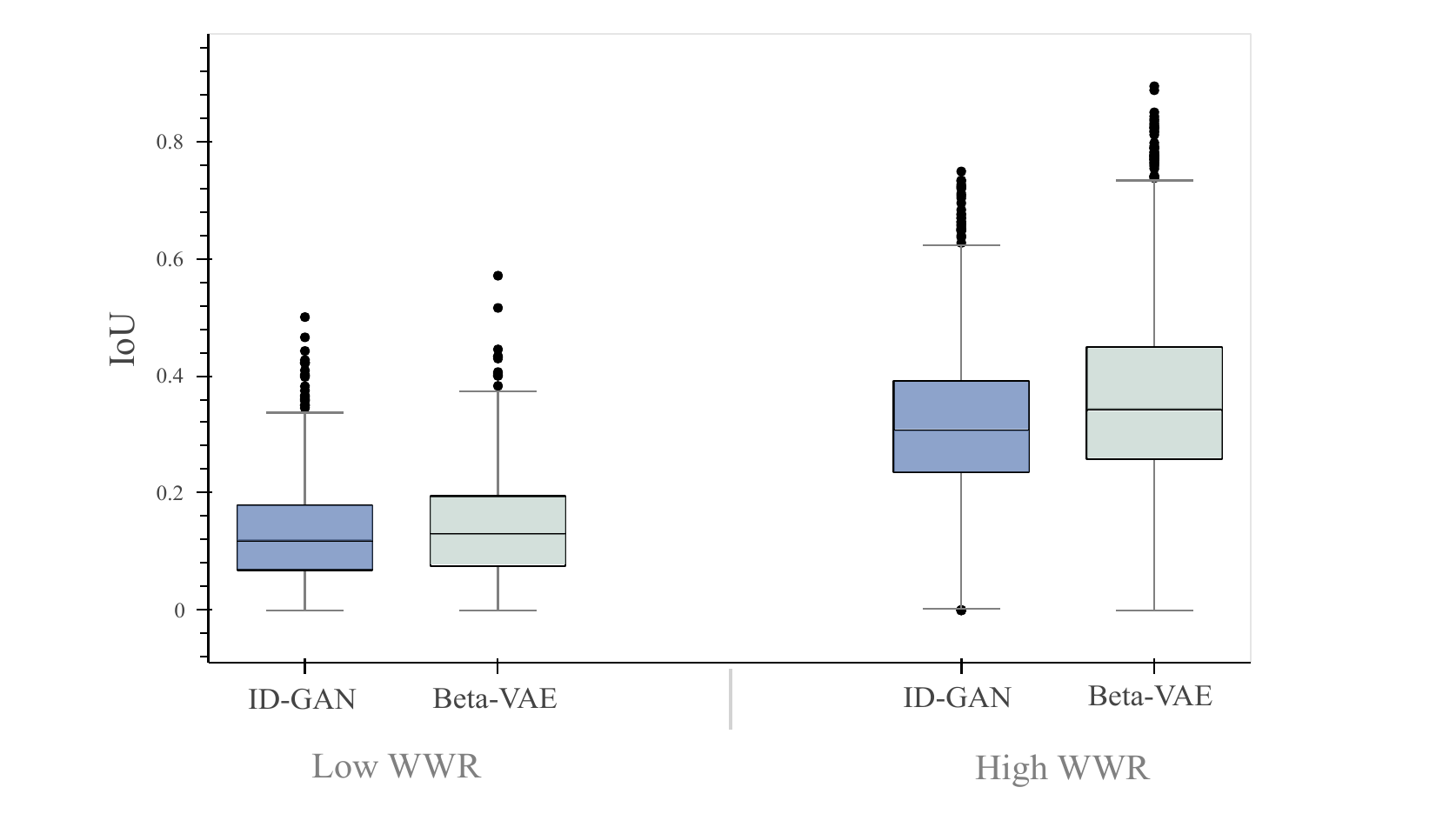}
    \caption{\label{fig:boxwwr}IoU distribution of the best performance samples in the ``Adding window''-task}
\end{figure}

To further address the unsatisfactory performance regarding WWR and examine its image-editing performance quantitatively with this respect, we designed a simple experiment. Specifically, we take out the image with zero WWR, i.e., images captured from the raw LOD1 models, from the triplet of each sensor-point, as shown in Fig. \ref{fig:dataset}. Then, extract their image representations via $\beta$-VAE encoder, and traverse the latent representations of WWR for 20 values, which will be fed into $ \beta$-VAE and ID-GAN to generate corresponding samples. Finally, we calculate the IoU of glazing for all the generated samples using images with low WWR and high WWR in the corresponding triplet as ground-truth, respectively. The IoU of the best-performing samples are recorded and plotted as the box plot in Fig. \ref{fig:boxwwr}. It can be observed that in the case of high WWR, which would exhibit more prominent impacts, both $\beta$-VAE and ID-GAN demonstrate acceptable performance with a large fraction of samples whose glazing IoU could exceed 0.4.

Thus, it could be argued that although $\beta$-VAE is not able to fully recover the details of window patterns in ground-truth, it is still able to capture some global and prominent characteristics related to WWR and guide ID-GAN to generate realistic samples for better interaction with the user. It is possible to confirm that SolarGAN has the potential for parametric study on WWR.

\subsection{Performance evaluation on time series}
\label{sec:eval_timeseries}

\begin{figure*}[htbp]
	\centering
	\includegraphics[width=1\textwidth]{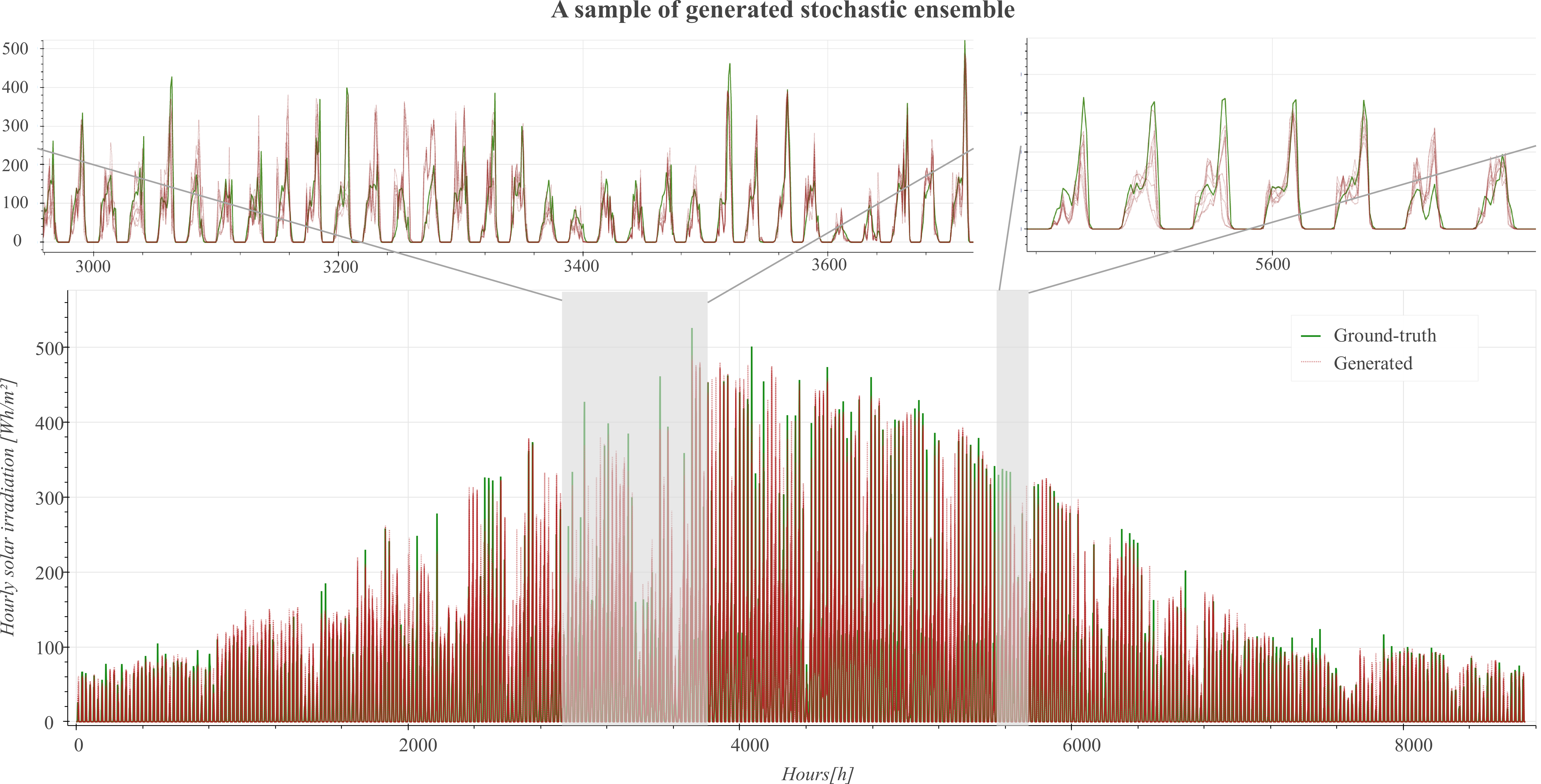}
    \caption{\label{fig:annual_sample}A typical sample of generated ensemble of annual hourly solar irradiance time-series in Zurich, with local zoom in on a monthly segment and weekly segment, in comparison to the ground-truth.}
\end{figure*}

\subsubsection{Statistical metrics to evaluate fidelity of the generated time series samples on a population basis}
\label{sec:timeseries_statistical}

A typical generated ensemble of annual hourly solar irradiance time-series on a sensor-point is shown in Fig. \ref{fig:annual_sample} It can be observed that, each individual sample within the generated ensemble differs in detail from the ground-truth, but the trend of the whole ensemble is consistent. For annual solar irradiance, we note that it could be characterized by 3 main aspects, including value distribution of the solar irradiations under the hourly and annual aggregation levels, short-term trend, i.e. hourly variation within a daily range, as well as long-term trend, i.e. daily variation on an annual basis. There are already some commonly-chosen statistical metrics to assess the fidelity of the samples on a population basis with these three aspects. Since we trained 2 conditional time-series generation models on the Singapore and the Zurich dataset, and each dataset contains samples using weather data from 5 Southeast Asian or European cities, here we only present the plots for the Singapore and Zurich cases, respectively. We will still compare statistical metrics for all cities in Table \ref{tab:site}. It is necessary to note that we obtain generated samples in ensemble generation, with an ensemble size of 10.

\begin{figure*}[htbp]
    \centering
	\includegraphics[width=1\textwidth]{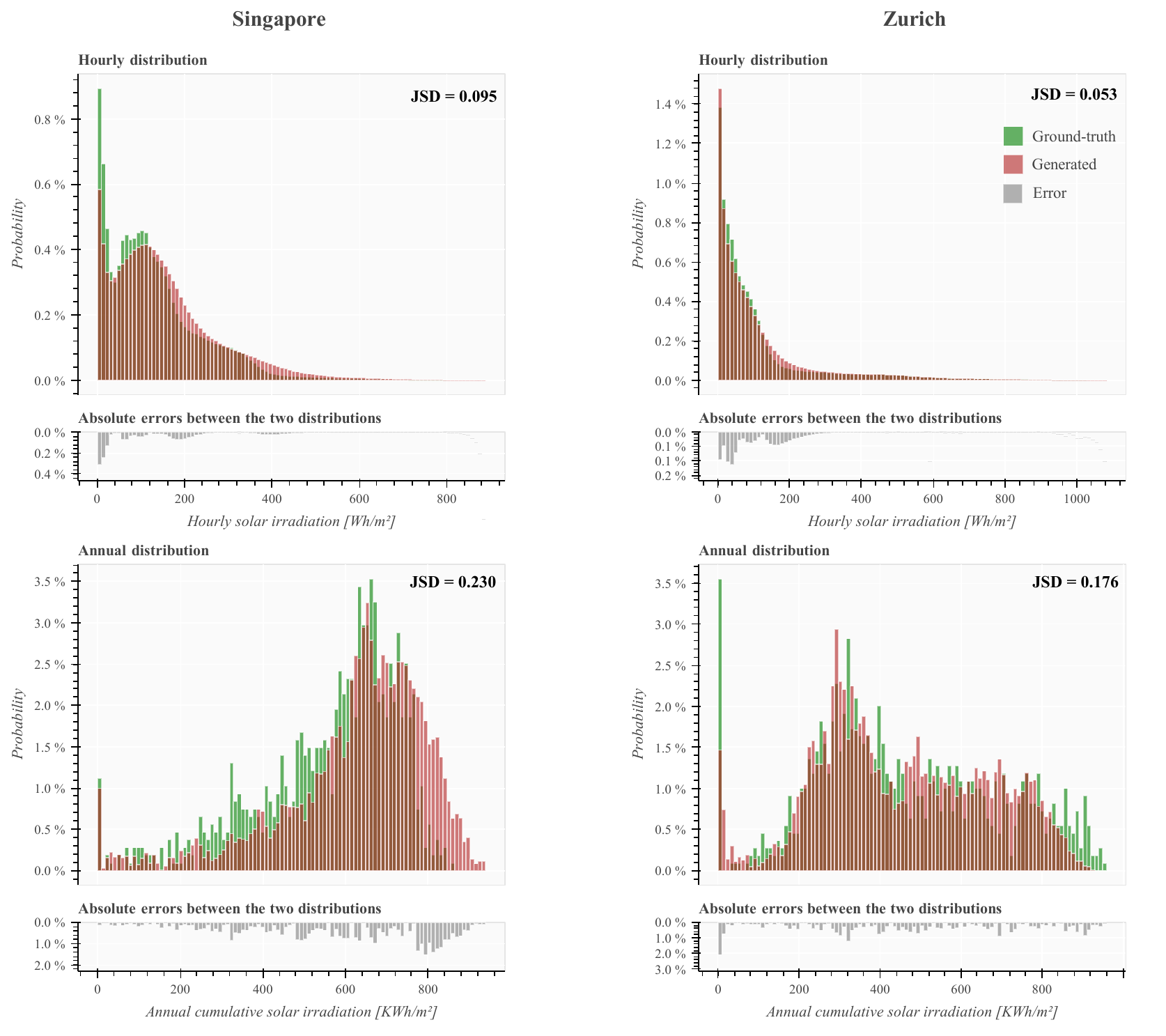} 
    \caption{\label{fig:sg_jsd}Histograms of hourly and annual solar irradiation value distributions of the ground-truth and generated data for the Singapore and Zurich case, respectively, with JSD between both distributions labeled and absolute errors between the ground-truth and generated distributions plotted right below}
\end{figure*}

First, regarding the value distribution. We sort all the values regardless of the time steps into 50 bins to form their respective histogram distributions for both generated as well as ground-truth samples. Then, similar to \cite{baasch2021conditional, dong2022data, lin2020using}, we apply Jensen-Shannon Divergence (JSD), as shown in Eq. \ref{eq_jsd}) to measure the similarity between these two histogram distributions. JSD is a variant from the classical statistical metric, Kullback–Leibler Divergence (KLD), $D_{K L}(P_{r} \| P_{g})=\sum_{x} P_{r}(x) \log \frac{P_{r}(x)}{P_{g}(x)}\label{eq_kld}$. Different from KLD, JSD is symmetric and thus better-suited for evaluation. As a rule of thumb, where the JSD is less than or equal to 0.1, the two distributions are generally considered very similar to each other\cite{baasch2021conditional}. Instead of only examining the raw hourly data, we check the similarity of distributions after aggregation on both the hourly and annual basis, respectively.

\begin{equation}
    \begin{aligned}
    D_{\mathrm{JS}}(P_{r} \| P_{g}) = & D_{\mathrm{KL}}\left(P_{\text {r}} \|\left(\frac{P_{\text {r }}+P_{g}}{2}\right)\right)+\\
    & D_{\mathrm{KL}}\left(P_{\mathrm{g}} \|\left(\frac{P_{\text {r }}+P_{g}}{2}\right)\right) \label{eq_jsd}
    \end{aligned}
\end{equation}

The histograms of the generated data and ground-truth for Singapore and Zurich are shown in Fig. \ref{fig:sg_jsd}. The absolute error between the ground-truth and the generated distributions (in percentage) specific to each bins has also been plotted right below, with JSD between the both distributions labeled, It is observed that in both cases, the generated data are close to the ground-truth on an hourly with JSD always around 0.1. However, on an annual basis, the generated samples exhibit a larger deviation from the ground-truth, which perhaps implies that generation by weekly patch, although ensuring more consistent details, might compromise the consistency of the annual cumulative data. In addition, the quality of the generated samples in the Zurich case is generally better than that in the Singapore case, as the values generated samples in Singapore are biased to higher ranges. This bias is actually due to the 
larger variation within the five Southeast Asian cities compared to the five European cities, which will be analysed in detail later referring to Table \ref{tab:site}. Nevertheless, it can also be observed that the Zurich case contains a large proportion of sensor-points subject to very low annual solar irradiation, which are not well captured by our model. Considering such sensor-points are obviously not of great BIPV potential, which means the setup of the simulation ground-truth dataset should also be refined. 

\begin{figure*}[htbp]
    \centering
	\includegraphics[width=1\textwidth]{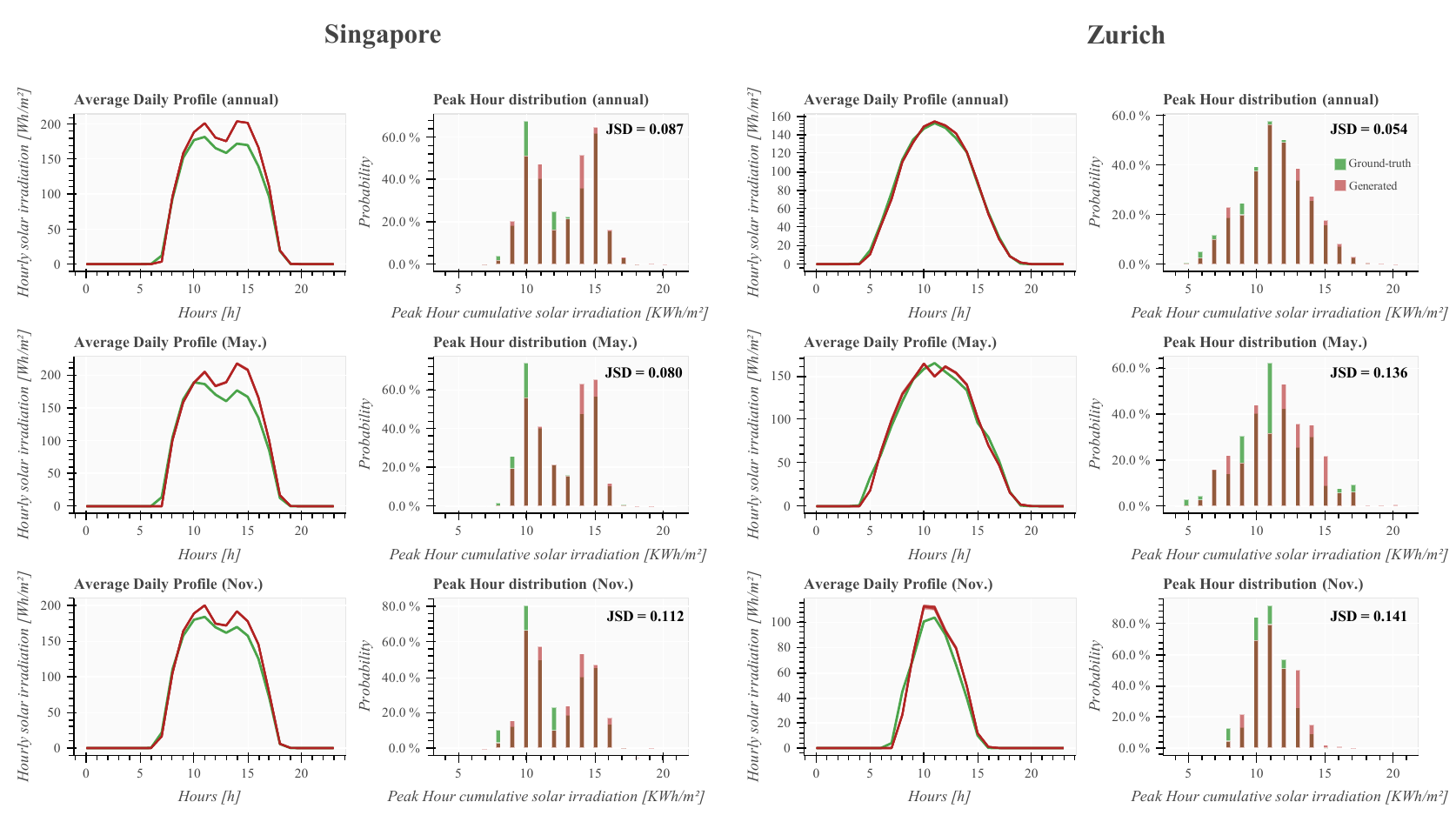}
    \caption{\label{fig:sg_ph}Annual and monthly (in May. and Nov.) solar irradiance average daily profiles and histogram of the daily peak hour distributions of the ground-truth and generated data for the Singapore case, with JSD between both distributions labeled}
\end{figure*}

Next, we need to ensure that the generated samples are also consistent with ground-truth in terms of temporal trend. For the short-term temporal trend, the daily peak hour of solar irradiance is of great interest to us. We could still make histograms of the peak hour distribution for generated and ground-truth samples, and still use JSD to measure the similarity between them. In addition, we also plot the daily solar profile that is averaged over all the days throughout the year and across all the samples to visually compare the pattern of variations between the generated and the ground-truth data.

The peak hour distributions and average daily profiles for Singapore and Zurich, as well as the respective JSD between the generated and ground-truth peak hour distributions, are shown in Fig. \ref{fig:sg_ph}. Here, we also add two typical months for each case for a more comprehensive evaluation. It could be observed that the peak hour distributions of generated data for both Singapore and Zurich are quite consistent with the ground-truth, the patterns of daily profiles are also well-captured by the generated data.

\begin{figure*}[htbp]
    \centering
	\includegraphics[width=1\textwidth]{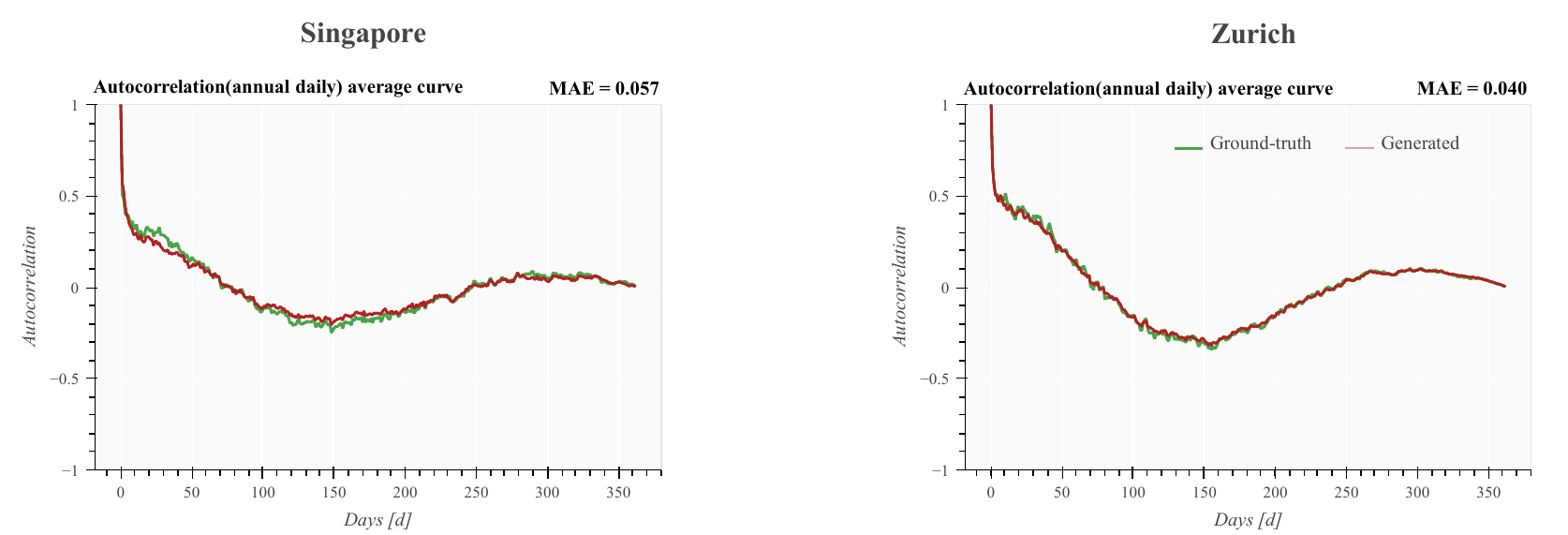}
    \caption{\label{fig:sg_zh_ac}Averaged annual auto-correlation curves of the daily cumulative solar irradiation for the Singapore (left) and Zurich case (right)}
\end{figure*}

Finally, we use annual autocorrelation curves on daily cumulative solar irradiation to compare the consistency of the generated samples with the ground-truth in terms of the long-term temporal trend. The definition of autocorrelation $r_k$, with $k$ being the time lag, is given in Eq. \ref{eq_ac}. 

\begin{equation}
   r_{k}=\frac{\sum_{t=1}^{T-k}\left(x_{t}-\bar{x}\right)\left(x_{t-k}-\bar{x}\right)}{\sum_{t=1}^{T}\left(x_{t}-\bar{x}\right)^{2}}\label{eq_ac}
\end{equation}

where, $T = 365$ in our case. It can be understood as calculating the covariance between a time series after shifting $k$ time-steps and itself. Increasing $k$ from 1 to $T-1$ on the year-long daily cumulative solar irradiation gives us an annual autocorrelation curve, which is a classic indicator of the long-term trend of the solar irradiation \cite{bartoli1981autocorrelation, dong2022data, lin2020using}. We use Mean Absolute Error (MAE) to calculate the discrepancy between the autocorrelation curves of all the generated samples and that of the ground-truth.

The annual daily autocorrelation curves averaged all the samples for Singapore and Zurich, as well as the corresponding MAE between the generated and ground-truth are shown in Fig. \ref{fig:sg_zh_ac}. Essentially, the generated samples in both cases has quite consistent trends with that of ground-truth data.

\begin{table*}[htbp]
\caption{Statistical validation results on the fidelity of the generated time-series samples using weather files from 5 European / Southeast Asian cities}
\label{tab:site}
\begin{tabular}{c|cc|cc}
\hline
\multirow{2}{*}{City} & \multicolumn{2}{c|}{Value distribution} & \multicolumn{2}{c}{Temporal trends} \\ \cline{2-5} 
                      & Hourly-JSD         & Annual-JSD         & Peak hour-JSD        & AC-MAE       \\ \hline
Geneva                & 0.047              & 0.161              & 0.120                & 0.044        \\
Milan                 & 0.040              & 0.152              & 0.083                & 0.040        \\
Paris                 & 0.048              & 0.153              & 0.071                & 0.041        \\
Berlin                & 0.044              & 0.173              & 0.070                & 0.041        \\
Zurich                & 0.053              & 0.176              & 0.054                & 0.040        \\ \hline
Ho Chi Minh City      & 0.113              & 0.221              & 0.056                & 0.063        \\
Jakarta               & 0.122              & 0.245              & 0.059                & 0.065        \\
Kuala Lumpur          & 0.132              & 0.257              & 0.077                & 0.081        \\
Bangkok               & 0.095              & 0.241              & 0.051                & 0.060        \\
Singapore             & 0.095              & 0.230              & 0.087                & 0.057        \\ \hline
\end{tabular}
\end{table*}

Table \ref{tab:site} summarizes the statistical validation results. Within the European or Southeast Asian cities, the results do not show obvious variation across cities, suggesting that one model is able to account for several cities as long as they share similar weather characteristics. 

Nevertheless, regarding the value distribution, the quality of the generated samples in the Southeast Asian case is generally worse than that in the European case. We speculate that the main reason for this difference is the higher inter-city variation in the range of solar irradiation among Southeast Asian cities (as demonstrated in Fig. \ref{fig:dataset}). Therefore, the model trained on the samples from the Southeast Asian cities fails to accurately capture the exact conditions specific to different cities, leading to bias. Whereas for the cases of European cities, such an issue could be alleviated due to the much smaller inter-city variation. This observation indicates that a more principled study on the proper geographical scope that one model should cover need to be conducted in the future. 

Regarding the short-term and long-term temporal trends, both cases seem to be satisfactory. The Southeast Asian case outperforms a bit better in terms of the short-term trend, possibly the daily solar patterns vary insignificantly with the seasons due to the year-long high solar altitude there. However, the seasonal variation patterns across these cities, although generally not very prominent, still somehow vary from each other. This probably causes the long-term trend performance falls behind the European case. 

\subsubsection{Classification modelling to evaluate time-series fidelity on different types of urban scenarios}
\label{sec:mlp_classification}

Considering that the generated data might serve as inputs to other downstream data-driven or machine-learning models, another alternative to evaluate the fidelity of generated data is to test whether the generated data and ground-truth are interchangeable for a particular machine-learning model. In general, we take the case where the model is trained and tested using only the generated data or ground-truth as the benchmark to reflect the capability of the model, i.e.,

\begin{itemize}
    \item Train on Real, Test on Real (TRTR)
    \item Train on Synthetic, Test on Synthetic (TSTS)
\end{itemize}

And the interchangeability between generated data and ground-truth is involved in the following scenarios,

\begin{itemize}
    \item Train on Real, Test on Synthetic (TRTS)
    \item Train on Synthetic, Test on Real (TSTR)
\end{itemize}

Specifically, we would like to train a time-series classifier to classify time-series samples according to different groups of urban context scenarios. We hope this design help us understand the respective fidelity of the generated samples corresponding to different types of urban context scenarios.

Therefore, we roughly divided the test-points in the test set 8 classes based on four orientations and two obstruction levels. Here, the obstruction level is defined by calculating the ratio of pixels occupied by the sky to the entire circular area in the fisheye image.  For test-points on facades, the maximum ratio is 0.5 and the minimum is 0. We consider a value between 0 and 0.25 as the high obstruction level class and a value between 0.25 and 0.5 as the low obstruction level class. We selected 1000 sensor-points in both Singapore and Zurich test datasets. The amount of low obstruction level test-points vs. that of high obstruction level is about 2:1. 

We selected weekly samples in June and August as the ground-truth for Singapore and Zurich, respectively, because the solar irradiance is high at these times. The corresponding generation process is also in ensemble generation with an ensemble size of 10, but we averaged the ensembles to ensure that the amount of ground-truth and generated samples are equal.

 \begin{figure*}[htbp]
	\centering
	\includegraphics[width=1\textwidth]{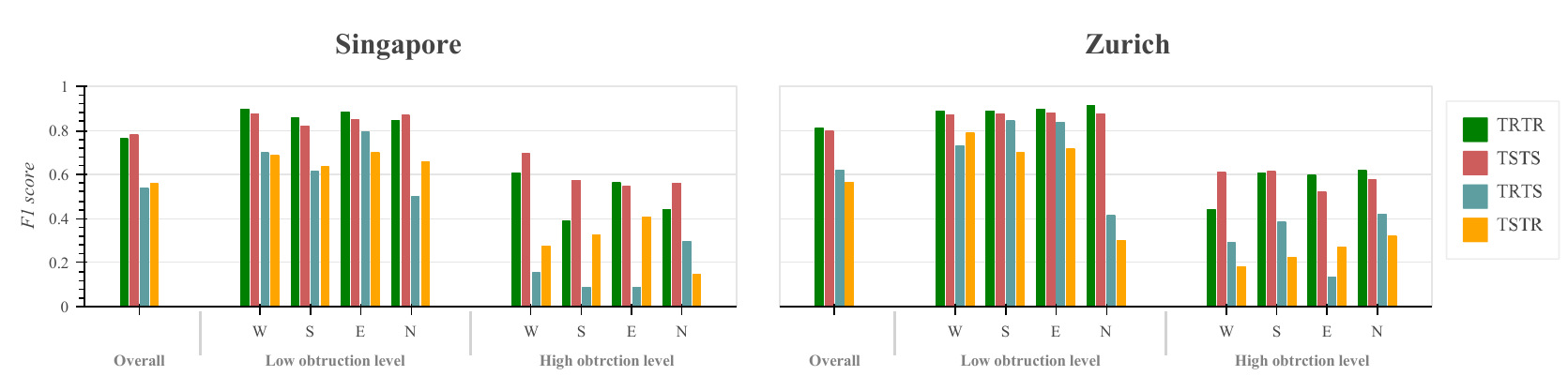}
    \caption{\label{fig:mlp}Performance comparison of the MLP classifier over 8 categories of the urban context scenarios. Singapore (left); Zurich (right)}
\end{figure*}

We built a time series classifier based on MLP. The model architecture is the same as the discriminator in the conditional time-series generator, as shown in Fig. \ref{fig:doppelganger}, except that a softmax layer is added to the last layer to output the classification results 8 categories. We train it for 500 epochs and use Cross-Entropy Loss as the loss function.

Finally, we choose the common evaluation metric F1 Score as the classification performance indicator. as shown in Eq. \ref{eq_f1} The terms in the formula are consistent with the classification results introduced in Section \ref{sec:eval_images}. The F1 Score expects the classifier to give accurate but not overly generous classification predictions. It lies between zero and one, the larger the better. We could also average the F1 Score over each class to reflect the overall classification performance of the model.

\begin{equation}
   F1\;Score=\frac{2 \# T P}{2\# T P+\# F P+\# F N} \label{eq_f1}
\end{equation}

We compare F1 Score of the MLP classifier for the Singapore and Zurich datasets in those aforementioned four cases in Fig. \ref{fig:mlp}. From the results of TSTS and TRTR cases, the classifier is able to achieve F1 scores around 0.75-0.8 on both Singapore and Zurich datasets, while the performance drops to below 0.6 for both the TSTR and TRTS cases, which is not that satisfying. However, if we look at the F1 scores of specific categories, we could see that the underperformance mainly lies in sensor-points with north orientation and high obstruction level, where the solar irradiance is usually low. It can be seen that the sensor-points with east, west, and south orientation and low obstruction level, which are our key interests in terms of solar potential, do not drop F1 Score significantly in the TRTS and TSTR cases. It is reasonable to say that the generated data reach acceptable interchangeability with the ground-truth.

\subsection{Quality of parametric output}
\label{sec:timeseries_parametric}

 \begin{figure*}[htbp]
	\centering
	\includegraphics[width=1\textwidth]{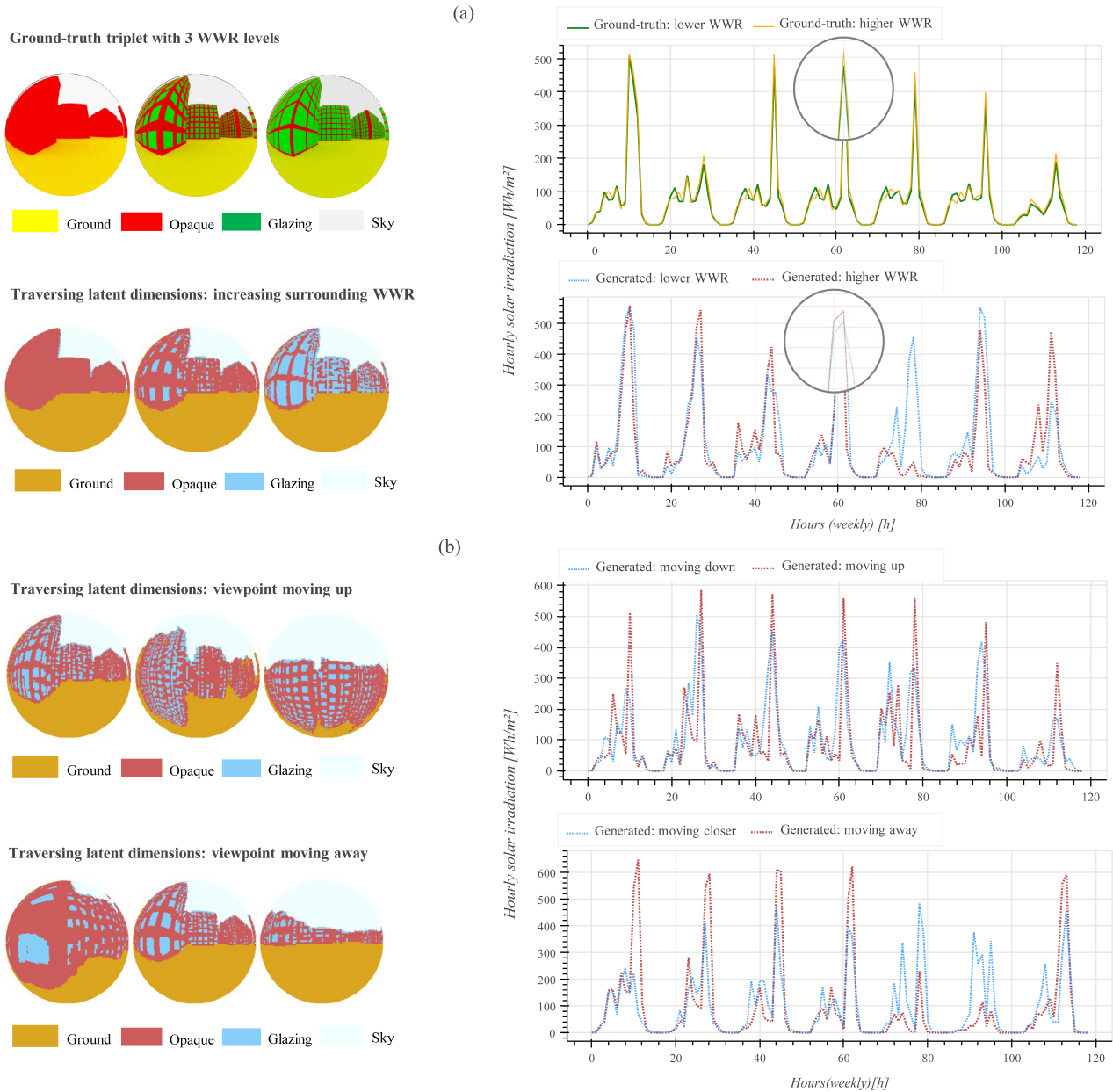}
    \caption{\label{fig:parametric}Parametrically changing urban context factors via fisheye image editing and the corresponding solar time-series outputs, taking one test-point as an example. For clarity, the generated images are recoloured and only one-week patches of the solar time-series under 2 extreme parametric inputs are plotted with ensemble size setting to 1. (a). Traversing in the latent dimension of WWR with zoom-in during one peak period, ground-truth vs. generated. (b) Traversing latent dimensions related to vertical (above) and horizontal (below) viewpoint shifting.}
\end{figure*}

We would like to further qualitatively exhibit our model's potential to conduct solar-driven parametric geometry design without 3D engines. First, WWR of the surrounding buildings is the intentionally considered parametric urban context variability when designing the dataset. As already evaluated in Section \ref{sec:eval_images}, the model is able to add windows and adjust WWR given a fisheye categorical mask, via traversing one of the latent representations, which means the model learnt to align the underlying variability of WWR in the dataset to one latent dimension without any supervision. Then, compared to the ground-truth solar time-series, as Fig. \ref{fig:parametric}(a) demonstrates, the time-series GAN is also able to generally capture the specific effects of changing the latent WWR in terms of magnitude and trend.

In addition, although not intentionally designed, our model could also unsupervisedly associate some latent dimensions with viewpoint coordinates, another underlying and prevailing variability in the dataset. As shown in Fig. \ref{fig:sg_wwr}(b), via traversing in specific latent dimensions, we are able to shift the view vertically and horizontally, as well as moving closer or further relative to the original sensor-point location . The time-series generator could also make reasonable responses. E.g., in overall generating larger solar irradiance data when moving above or away from the surrounding obstacles. Note that since now the model is generating images and time-series that do not exist in the dataset at all, currently there is no ground-truth for comparison.

Despite the appealing potential, both image editing and solar time-series generation have room for improvement in various details. Moreover, we still lack principled guidelines for dataset design and evaluation criteria selection, when considering the complicated impacts from these urban context factors. Therefore, current analysis is limited to qualitative observation on some samples, requiring further studies.

\subsection{Computing time}
\label{sec:computing_time}

 \begin{figure}[h]
	\begin{center}
	\includegraphics[width=0.45\textwidth]{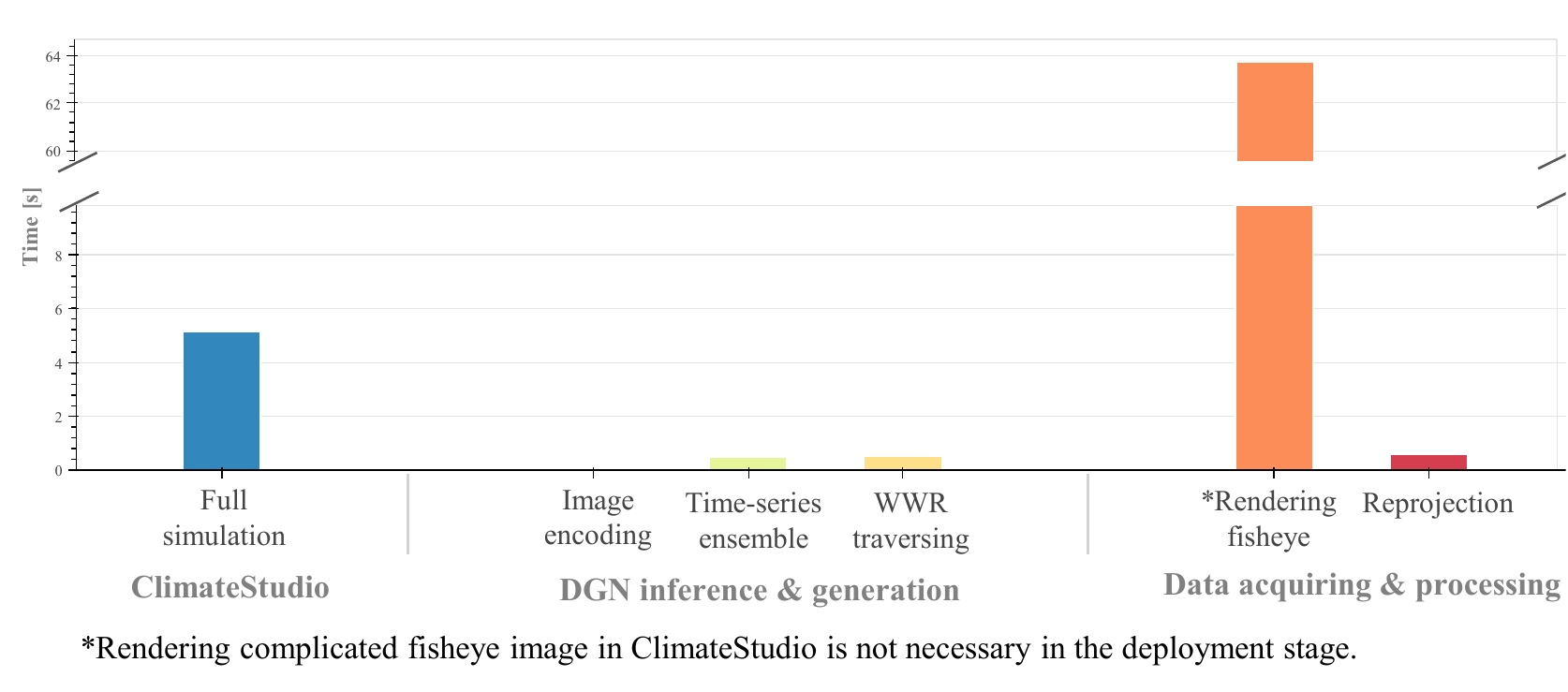}
    \caption{\label{fig:computing_time}Computing time Comparison between ClimateStudio and SolarGAN }
  	\end{center}
\end{figure}

We also conduct a simple test on the computing time, as speed is one of the major advantage of data-driven approaches. Specifically, we sample 10 new sensor-points scattered over the LOD1 model and repeat all the necessary workflow in our model to generate solar time-series ensemble and modifying context WWR, as shown in Fig. \ref{fig:computing_time}. We also let ClimateStudio simulate solar irradiance for these 10 sensor-points for comparison. 

As can be seen, although ClimateStudio is already quite fast, all the operations in DGNs are almost real-time, less than one second, even though we enable ensemble time-series generation as well as traversing several times in the latent dimension of WWR. However, obviously, capturing images is indeed the bottleneck that impedes the speed of the entire workflow, since it is rendered by ClimateStudio. For now, by doing so we could conveniently get images and solar time-series in the same environment. However, it is quite unnecessary to use rendering engine to get such simple images as categorical masks. In the Discussion, we will elaborate on possible alternatives to capture images.

\section{Discussion}
\label{sec:discussion}

We propose a model architecture as a high resolution urban solar irradiance time-series generator. By validating the fidelity of the generated samples, we could confirm that the proposed model based on DGNs and open-source urban geometry GIS data is a feasible and promising solution as a fast surrogate model for time-consuming simulation engines. Unlike other data-driven models that usually compromise on resolution, the spatial-temporal resolution of our model is consistent with that of the simulation engine, thanks to the apt input forms of urban context as fisheye categorical mask images inspired by previous urban solar studies, as well as appropriate DGN model architectures following the state-of-the-art deep-learning research. Our model’s capability of image editing and conditional time-series generation could also potentially allow for parametric design on several urban context factors without 3D engines. 

The proposed model could be potentially used as a lightweight plug-in with CAD or GIS software for fast solar irradiance prediction. It could also be combined with advanced urban energy modeling software with 3D capabilities to provide stochastic boundary conditions. What we also wish to highlight is its potential for direct handling of real-world images, thanks to the rapid development of urban scenario segmentation applications in recent years \cite{szczesniak2022method}. The label mask obtained from the semantic segmentation is highly similar to what we need. Such urban scenario label masks could also inform us of installation feasibility on specific locations. This means that we could obtain the annual PV yield profile for arbitrary urban locations as long as some photos are provided. 

However, as a preliminary study, we also found that almost each part of the current workflow reveals many problems that need to be studied systematically in the future:

\begin{itemize}
    \item Regarding the dataset setup, we need to improve the comprehensiveness of potential variabilities reflected in the dataset, especially for urban context factors except for WWR, such as surface inclination, vegetation and tree, surface albedo settings. This is decisive for the generalization performance of the model and even more significant than the model architecture design. Therefore, we still need to follow closely the advances in physics-based urban PV studies, especially sensitivity and uncertainty analysis based on simulations and real-world experiments.
    
    \item Regarding model architecture, although we propose a feasible architecture, it is probably not yet optimal. We need to conduct a more exhaustive benchmark study to select appropriate backbone network architectures and determine some important prior settings and hyperparameters, e.g., the proper geographical scope one model should cover, time-series patch length in generation, dimension of the latent image representation, etc. We also need to qualitatively evaluate the model’s response to parametric input of urban context factors when a more comprehensive dataset is available. In addition, 
    
    \item Regarding the acquisition of data, especially the way we capture fisheye categorical mask images, although this is not the focus of this paper, using ClimateStudio to render such simple images is an obvious overkill, which significantly impedes the overall speed. In the future, alternative approaches similar to \cite{liang2020solar3d, middel2017sky}, which are based on taking real-time screenshots in the GIS platform, need to be investigated. Even so, reading, writing, and transferring images might be inevitably rather time-consuming in practice. To mitigate this issue, we could possibly leverage the model's capabilities of extracting image representations and editing images. Specifically, some representative sensor-points could be selected beforehand, with their corresponding images captured and fed into the model. Then, the model yields their low-dimensional image representations and only saves them as GIS data instead of saving a lot of images.  For other sensor-points in adjacent regions to those representative ones, their image representations could be obtained via interpolating the latent dimensions associated with viewpoint coordinates. This is similar to the parametric design workflow shown in Fig. \ref{fig:parametric} but will be used as a special 3D spatial interpolation method.
\end{itemize}

\section{Conclusion}
\label{sec:conclusion}

In this work we present a methodology using DGN to generate realistic urban solar profiles with easily accessible data inputs, namely, simple fisheye images captured from LOD1 model available from open-source GIS databases. The current model is trained using simulated data from accurate physics-based solar simulation engine, and we details the entire workflow of preparing simulation dataset and data pre-processing techniques. We also elaborate on the rationale and procedure to design model architecture based on established work. Using geometry and weather data from both European and Southeast Asian cities, we demonstrate that as a surrogate model to the simulation engine, it is able to generate results with satisfactory fidelity. It achieves sufficient spatial-temporal resolution as simulated results while being able to introduce variability to the deterministic simulation results.  In addition, it even exhibits great potential for parametric study. Thus, it can be efficiently used on a large-scale urban solar potential evaluation with high resolution. Combining with other established urban scenario segmentation applications, it may also be fed with real-world imagery with only minor modification. 

\appendix
\section{Appendix}

 \begin{figure*}
	\begin{center}
	\includegraphics[width=1\textwidth]{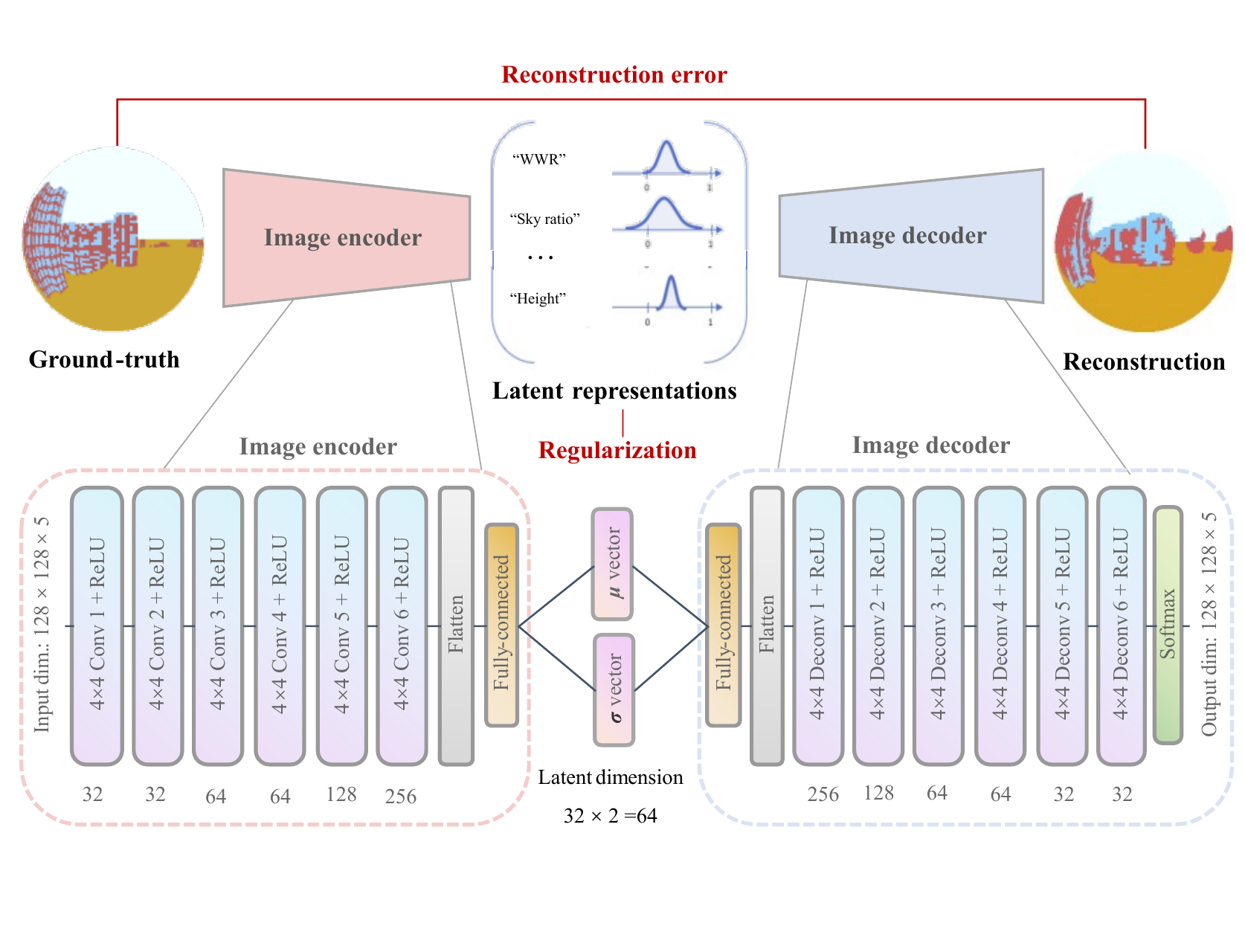}
    \caption{\label{fig:betavae} Detailed network architecture of the image representation-learning network based on $\beta$-VAE \cite{lee2020high}, lines and texts in red indicate the loss function terms involved. The value under the convolutional layers indicate the number of convolutional filters in this layer. The sample images has been recolored and shown in the fisheye format for more intuitive visual appearance, the actual input and output images are still one-hot encoding cube-maps}
  	\end{center}
\end{figure*}

 \begin{figure*}
	\begin{center}
	\includegraphics[width=1\textwidth]{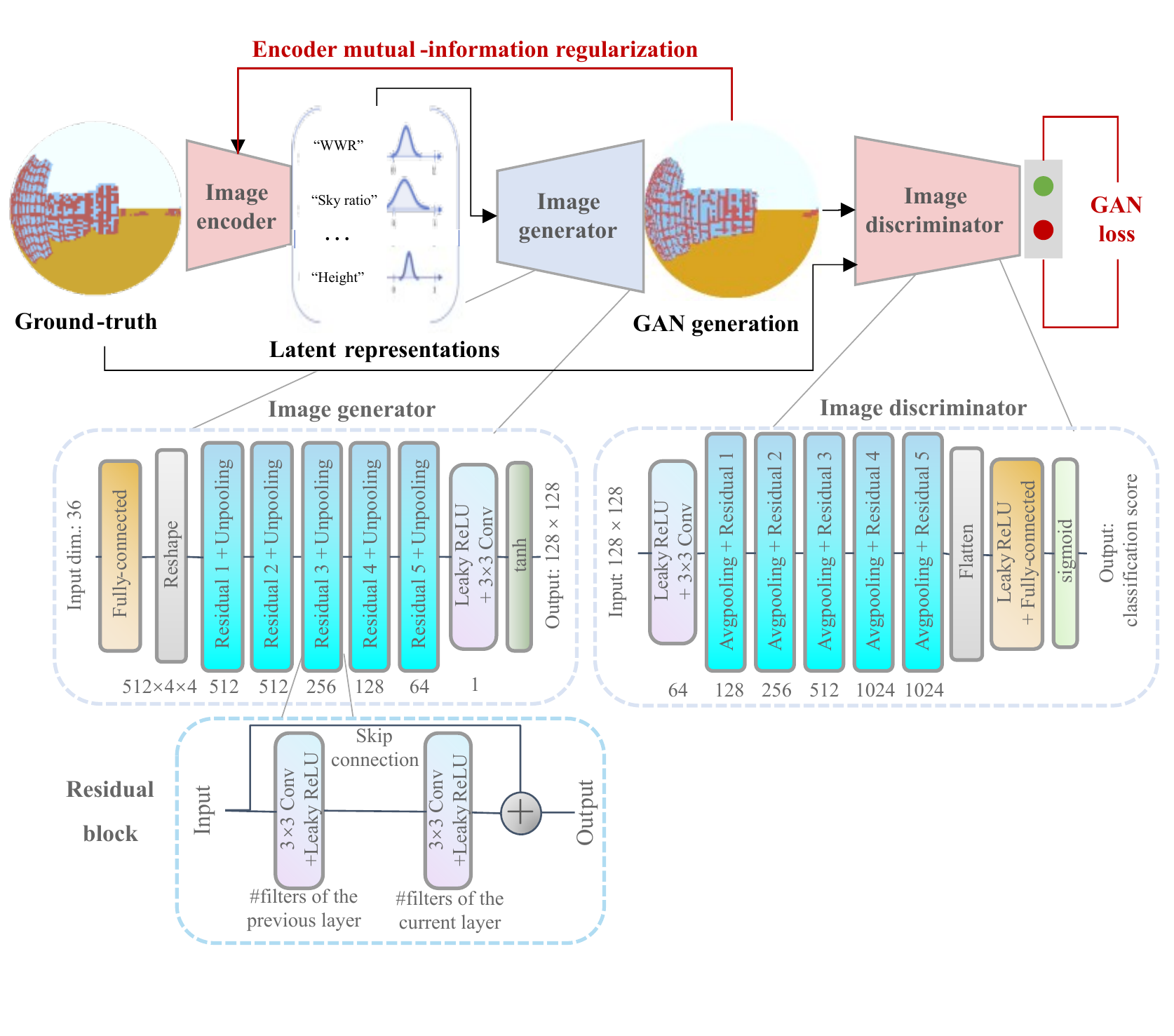}
    \caption{\label{fig:idgan} Detailed network architecture of the high-fidelity image editing network based on ID-GAN \cite{lee2020high}, lines and texts in red indicate the loss function terms involved. The value under the convolutional layers indicate the number of convolutional filters in this layer. The sample images has been recolored and shown in the fisheye format for more intuitive visual appearance, the actual input and output images are still order encoding grayscale cube-maps}
  	\end{center}
\end{figure*}

 \begin{figure*}
	\begin{center}
	\includegraphics[width=1\textwidth]{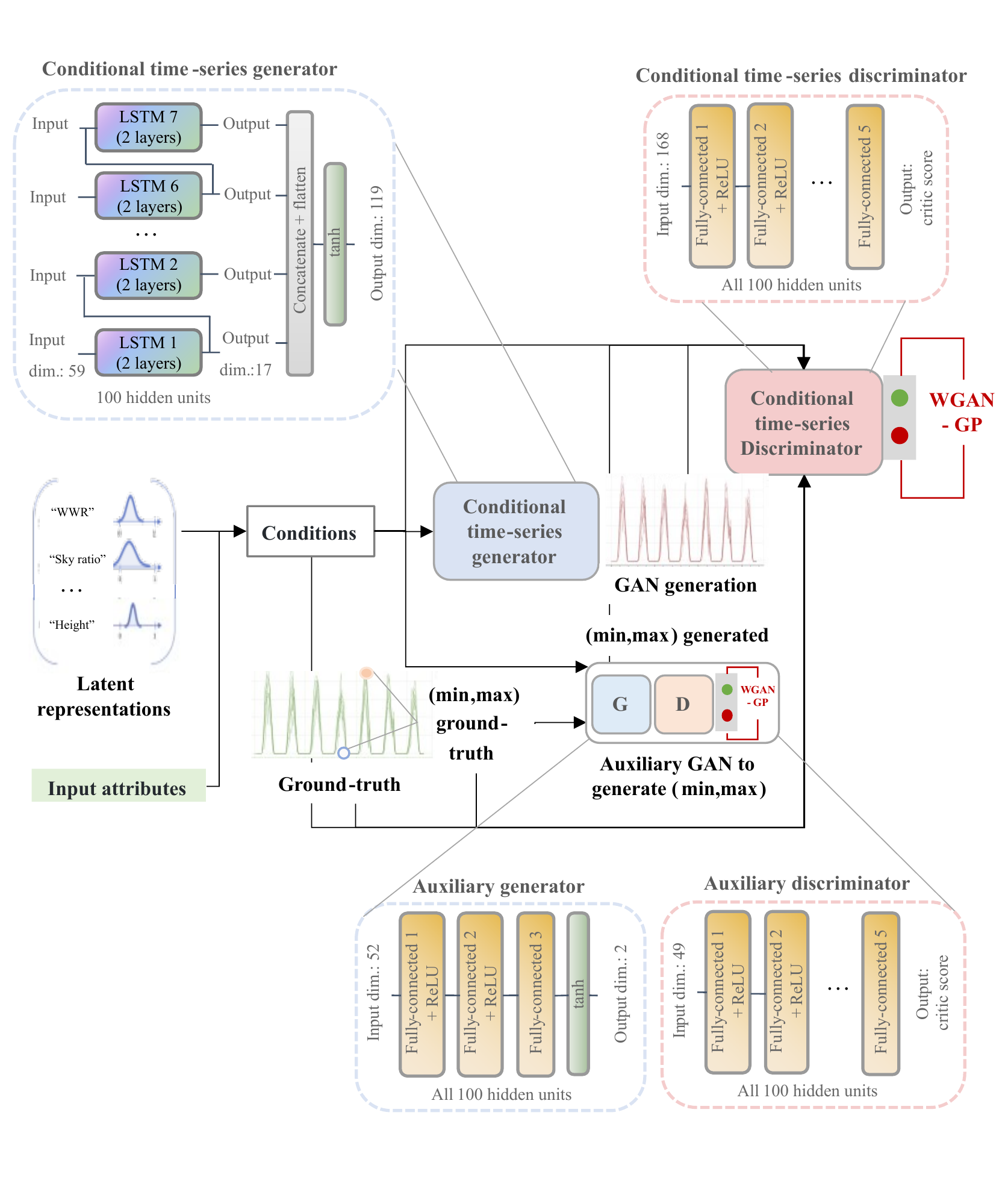}
    \caption{\label{fig:doppelganger} Detailed network architecture of the conditional time-series GAN based on DopperlGANger \cite{lin2020using}, lines and texts in red indicate the loss function terms involved. For LSTM cells, the number of hidden units denotes the output dimension of the MLP layer within it}
  	\end{center}
\end{figure*}


\printcredits

\bibliographystyle{model1-num-names}

\bibliography{cas-refs}

\begin{thebibliography}{43}
\expandafter\ifx\csname natexlab\endcsname\relax\def\natexlab#1{#1}\fi
\providecommand{\url}[1]{\texttt{#1}}
\providecommand{\href}[2]{#2}
\providecommand{\path}[1]{#1}
\providecommand{\DOIprefix}{doi:}
\providecommand{\ArXivprefix}{arXiv:}
\providecommand{\URLprefix}{URL: }
\providecommand{\Pubmedprefix}{pmid:}
\providecommand{\doi}[1]{\href{http://dx.doi.org/#1}{\path{#1}}}
\providecommand{\Pubmed}[1]{\href{pmid:#1}{\path{#1}}}
\providecommand{\bibinfo}[2]{#2}
\ifx\xfnm\relax \def\xfnm[#1]{\unskip,\space#1}\fi
\bibitem[{Izquierdo et~al.(2008)Izquierdo, Rodrigues, and
  Fueyo}]{izquierdo2008method}
\bibinfo{author}{S.~Izquierdo}, \bibinfo{author}{M.~Rodrigues},
  \bibinfo{author}{N.~Fueyo},
\newblock \bibinfo{title}{A method for estimating the geographical distribution
  of the available roof surface area for large-scale photovoltaic
  energy-potential evaluations},
\newblock \bibinfo{journal}{Solar Energy} \bibinfo{volume}{82}
  (\bibinfo{year}{2008}) \bibinfo{pages}{929--939}.
\bibitem[{Ward and Shakespeare(1998)}]{ward1998rendering}
\bibinfo{author}{G.~Ward}, \bibinfo{author}{R.~Shakespeare},
\newblock \bibinfo{title}{Rendering with radiance: the art and science of
  lighting visualization}  (\bibinfo{year}{1998}).
\bibitem[{Reinhart et~al.(2006)}]{reinhart2006tutorial}
\bibinfo{author}{C.~F. Reinhart}, et~al.,
\newblock \bibinfo{title}{Tutorial on the use of daysim simulations for
  sustainable design}  (\bibinfo{year}{2006}).
\bibitem[{Waibel et~al.(2017)Waibel, Evins, and
  Carmeliet}]{waibel2017efficient}
\bibinfo{author}{C.~Waibel}, \bibinfo{author}{R.~Evins},
  \bibinfo{author}{J.~Carmeliet},
\newblock \bibinfo{title}{Efficient time-resolved 3d solar potential
  modelling},
\newblock \bibinfo{journal}{Solar Energy} \bibinfo{volume}{158}
  (\bibinfo{year}{2017}) \bibinfo{pages}{960--976}.
\bibitem[{LLC(2021)}]{climatestudio}
\bibinfo{author}{S.~LLC}, \bibinfo{title}{Climatestudio}, \bibinfo{year}{2021}.
  \URLprefix \url{https://www.solemma.com/climatestudio}.
\bibitem[{Oke et~al.(2017)Oke, Mills, Christen, and Voogt}]{oke2017urban}
\bibinfo{author}{T.~R. Oke}, \bibinfo{author}{G.~Mills},
  \bibinfo{author}{A.~Christen}, \bibinfo{author}{J.~A. Voogt},
  \bibinfo{title}{Urban climates}, \bibinfo{publisher}{Cambridge University
  Press}, \bibinfo{year}{2017}.
\bibitem[{Liang et~al.(2020)Liang, Gong, Xie, and Sun}]{liang2020solar3d}
\bibinfo{author}{J.~Liang}, \bibinfo{author}{J.~Gong},
  \bibinfo{author}{X.~Xie}, \bibinfo{author}{J.~Sun},
\newblock \bibinfo{title}{Solar3d: A 3d extension of grass gis r. sun for
  estimating solar radiation in urban environments}  (\bibinfo{year}{2020}).
\bibitem[{Middel et~al.(2017)Middel, Lukasczyk, Maciejewski
  et~al.}]{middel2017sky}
\bibinfo{author}{A.~Middel}, \bibinfo{author}{J.~Lukasczyk},
  \bibinfo{author}{R.~Maciejewski}, et~al.,
\newblock \bibinfo{title}{Sky view factors from synthetic fisheye photos for
  thermal comfort routing—a case study in phoenix, arizona},
\newblock \bibinfo{journal}{Urban Planning} \bibinfo{volume}{2}
  (\bibinfo{year}{2017}) \bibinfo{pages}{19--30}.
\bibitem[{Peronato et~al.(2018)Peronato, Rastogi, Rey, and
  Andersen}]{peronato2018toolkit}
\bibinfo{author}{G.~Peronato}, \bibinfo{author}{P.~Rastogi},
  \bibinfo{author}{E.~Rey}, \bibinfo{author}{M.~Andersen},
\newblock \bibinfo{title}{A toolkit for multi-scale mapping of the solar
  energy-generation potential of buildings in urban environments under
  uncertainty},
\newblock \bibinfo{journal}{Solar Energy} \bibinfo{volume}{173}
  (\bibinfo{year}{2018}) \bibinfo{pages}{861--874}.
\bibitem[{Assouline et~al.(2017)Assouline, Mohajeri, and
  Scartezzini}]{assouline2017quantifying}
\bibinfo{author}{D.~Assouline}, \bibinfo{author}{N.~Mohajeri},
  \bibinfo{author}{J.-L. Scartezzini},
\newblock \bibinfo{title}{Quantifying rooftop photovoltaic solar energy
  potential: A machine learning approach},
\newblock \bibinfo{journal}{Solar Energy} \bibinfo{volume}{141}
  (\bibinfo{year}{2017}) \bibinfo{pages}{278--296}.
\bibitem[{Nault et~al.(2017)Nault, Moonen, Rey, and
  Andersen}]{nault2017predictive}
\bibinfo{author}{E.~Nault}, \bibinfo{author}{P.~Moonen},
  \bibinfo{author}{E.~Rey}, \bibinfo{author}{M.~Andersen},
\newblock \bibinfo{title}{Predictive models for assessing the passive solar and
  daylight potential of neighborhood designs: A comparative proof-of-concept
  study},
\newblock \bibinfo{journal}{Building and Environment} \bibinfo{volume}{116}
  (\bibinfo{year}{2017}) \bibinfo{pages}{1--16}.
\bibitem[{Walch et~al.(2020)Walch, Castello, Mohajeri, and
  Scartezzini}]{walch2020fast}
\bibinfo{author}{A.~Walch}, \bibinfo{author}{R.~Castello},
  \bibinfo{author}{N.~Mohajeri}, \bibinfo{author}{J.-L. Scartezzini},
\newblock \bibinfo{title}{A fast machine learning model for large-scale
  estimation of annual solar irradiation on rooftops},
\newblock in: \bibinfo{booktitle}{Proceedings of Solar World Congress 2019},
  \bibinfo{number}{CONF}, \bibinfo{organization}{International Solar Energy
  Society ISES}, \bibinfo{year}{2020}.
\bibitem[{Chen et~al.(2018)Chen, Wang, Kirschen, and Zhang}]{chen2018model}
\bibinfo{author}{Y.~Chen}, \bibinfo{author}{Y.~Wang},
  \bibinfo{author}{D.~Kirschen}, \bibinfo{author}{B.~Zhang},
\newblock \bibinfo{title}{Model-free renewable scenario generation using
  generative adversarial networks},
\newblock \bibinfo{journal}{IEEE Transactions on Power Systems}
  \bibinfo{volume}{33} (\bibinfo{year}{2018}) \bibinfo{pages}{3265--3275}.
\bibitem[{Fekri et~al.(2020)Fekri, Ghosh, and Grolinger}]{fekri2020generating}
\bibinfo{author}{M.~N. Fekri}, \bibinfo{author}{A.~M. Ghosh},
  \bibinfo{author}{K.~Grolinger},
\newblock \bibinfo{title}{Generating energy data for machine learning with
  recurrent generative adversarial networks},
\newblock \bibinfo{journal}{Energies} \bibinfo{volume}{13}
  (\bibinfo{year}{2020}) \bibinfo{pages}{130}.
\bibitem[{Zhang et~al.(2018)Zhang, Kuppannagari, Kannan, and
  Prasanna}]{zhang2018generative}
\bibinfo{author}{C.~Zhang}, \bibinfo{author}{S.~R. Kuppannagari},
  \bibinfo{author}{R.~Kannan}, \bibinfo{author}{V.~K. Prasanna},
\newblock \bibinfo{title}{Generative adversarial network for synthetic time
  series data generation in smart grids},
\newblock in: \bibinfo{booktitle}{2018 IEEE International Conference on
  Communications, Control, and Computing Technologies for Smart Grids
  (SmartGridComm)}, \bibinfo{organization}{IEEE}, \bibinfo{year}{2018}, pp.
  \bibinfo{pages}{1--6}.
\bibitem[{Khayatian et~al.(2021)Khayatian, Nagy, and
  Bollinger}]{khayatian2021using}
\bibinfo{author}{F.~Khayatian}, \bibinfo{author}{Z.~Nagy},
  \bibinfo{author}{A.~Bollinger},
\newblock \bibinfo{title}{Using generative adversarial networks to evaluate
  robustness of reinforcement learning agents against uncertainties},
\newblock \bibinfo{journal}{Energy and Buildings} \bibinfo{volume}{251}
  (\bibinfo{year}{2021}) \bibinfo{pages}{111334}.
\bibitem[{Baasch et~al.(2021)Baasch, Rousseau, and
  Evins}]{baasch2021conditional}
\bibinfo{author}{G.~Baasch}, \bibinfo{author}{G.~Rousseau},
  \bibinfo{author}{R.~Evins},
\newblock \bibinfo{title}{A conditional generative adversarial network for
  energy use in multiple buildings using scarce data},
\newblock \bibinfo{journal}{Energy and AI} \bibinfo{volume}{5}
  (\bibinfo{year}{2021}) \bibinfo{pages}{100087}.
\bibitem[{Dong et~al.(2022)Dong, Chen, and Yang}]{dong2022data}
\bibinfo{author}{W.~Dong}, \bibinfo{author}{X.~Chen},
  \bibinfo{author}{Q.~Yang},
\newblock \bibinfo{title}{Data-driven scenario generation of renewable energy
  production based on controllable generative adversarial networks with
  interpretability},
\newblock \bibinfo{journal}{Applied Energy} \bibinfo{volume}{308}
  (\bibinfo{year}{2022}) \bibinfo{pages}{118387}.
\bibitem[{Kingma and Welling(2013)}]{kingma2013auto}
\bibinfo{author}{D.~P. Kingma}, \bibinfo{author}{M.~Welling},
\newblock \bibinfo{title}{Auto-encoding variational bayes},
\newblock \bibinfo{journal}{arXiv preprint arXiv:1312.6114}
  (\bibinfo{year}{2013}).
\bibitem[{Goodfellow et~al.(2014)Goodfellow, Pouget-Abadie, Mirza, Xu,
  Warde-Farley, Ozair, Courville, and Bengio}]{goodfellow2014generative}
\bibinfo{author}{I.~Goodfellow}, \bibinfo{author}{J.~Pouget-Abadie},
  \bibinfo{author}{M.~Mirza}, \bibinfo{author}{B.~Xu},
  \bibinfo{author}{D.~Warde-Farley}, \bibinfo{author}{S.~Ozair},
  \bibinfo{author}{A.~Courville}, \bibinfo{author}{Y.~Bengio},
\newblock \bibinfo{title}{Generative adversarial nets},
\newblock \bibinfo{journal}{Advances in neural information processing systems}
  \bibinfo{volume}{27} (\bibinfo{year}{2014}).
\bibitem[{Chen et~al.(2016)Chen, Duan, Houthooft, Schulman, Sutskever, and
  Abbeel}]{chen2016infogan}
\bibinfo{author}{X.~Chen}, \bibinfo{author}{Y.~Duan},
  \bibinfo{author}{R.~Houthooft}, \bibinfo{author}{J.~Schulman},
  \bibinfo{author}{I.~Sutskever}, \bibinfo{author}{P.~Abbeel},
\newblock \bibinfo{title}{Infogan: Interpretable representation learning by
  information maximizing generative adversarial nets},
\newblock in: \bibinfo{booktitle}{Proceedings of the 30th International
  Conference on Neural Information Processing Systems}, \bibinfo{year}{2016},
  pp. \bibinfo{pages}{2180--2188}.
\bibitem[{Donahue et~al.(2016)Donahue, Kr{\"a}henb{\"u}hl, and
  Darrell}]{donahue2016adversarial}
\bibinfo{author}{J.~Donahue}, \bibinfo{author}{P.~Kr{\"a}henb{\"u}hl},
  \bibinfo{author}{T.~Darrell},
\newblock \bibinfo{title}{Adversarial feature learning},
\newblock \bibinfo{journal}{arXiv preprint arXiv:1605.09782}
  (\bibinfo{year}{2016}).
\bibitem[{Isola et~al.(2017)Isola, Zhu, Zhou, and Efros}]{isola2017image}
\bibinfo{author}{P.~Isola}, \bibinfo{author}{J.-Y. Zhu},
  \bibinfo{author}{T.~Zhou}, \bibinfo{author}{A.~A. Efros},
\newblock \bibinfo{title}{Image-to-image translation with conditional
  adversarial networks},
\newblock in: \bibinfo{booktitle}{Proceedings of the IEEE conference on
  computer vision and pattern recognition}, \bibinfo{year}{2017}, pp.
  \bibinfo{pages}{1125--1134}.
\bibitem[{Larsen et~al.(2016)Larsen, S{\o}nderby, Larochelle, and
  Winther}]{larsen2016autoencoding}
\bibinfo{author}{A.~B.~L. Larsen}, \bibinfo{author}{S.~K. S{\o}nderby},
  \bibinfo{author}{H.~Larochelle}, \bibinfo{author}{O.~Winther},
\newblock \bibinfo{title}{Autoencoding beyond pixels using a learned similarity
  metric},
\newblock in: \bibinfo{booktitle}{International conference on machine
  learning}, \bibinfo{organization}{PMLR}, \bibinfo{year}{2016}, pp.
  \bibinfo{pages}{1558--1566}.
\bibitem[{Zhu et~al.(2017)Zhu, Park, Isola, and Efros}]{zhu2017unpaired}
\bibinfo{author}{J.-Y. Zhu}, \bibinfo{author}{T.~Park},
  \bibinfo{author}{P.~Isola}, \bibinfo{author}{A.~A. Efros},
\newblock \bibinfo{title}{Unpaired image-to-image translation using
  cycle-consistent adversarial networks},
\newblock in: \bibinfo{booktitle}{Proceedings of the IEEE international
  conference on computer vision}, \bibinfo{year}{2017}, pp.
  \bibinfo{pages}{2223--2232}.
\bibitem[{Lee et~al.(2020)Lee, Kim, Hong, and Lee}]{lee2020high}
\bibinfo{author}{W.~Lee}, \bibinfo{author}{D.~Kim}, \bibinfo{author}{S.~Hong},
  \bibinfo{author}{H.~Lee},
\newblock \bibinfo{title}{High-fidelity synthesis with disentangled
  representation},
\newblock in: \bibinfo{booktitle}{European Conference on Computer Vision},
  \bibinfo{organization}{Springer}, \bibinfo{year}{2020}, pp.
  \bibinfo{pages}{157--174}.
\bibitem[{Hinton and Salakhutdinov(2006)}]{hinton2006reducing}
\bibinfo{author}{G.~E. Hinton}, \bibinfo{author}{R.~R. Salakhutdinov},
\newblock \bibinfo{title}{Reducing the dimensionality of data with neural
  networks},
\newblock \bibinfo{journal}{science} \bibinfo{volume}{313}
  (\bibinfo{year}{2006}) \bibinfo{pages}{504--507}.
\bibitem[{Higgins et~al.(2016)Higgins, Matthey, Pal, Burgess, Glorot,
  Botvinick, Mohamed, and Lerchner}]{higgins2016beta}
\bibinfo{author}{I.~Higgins}, \bibinfo{author}{L.~Matthey},
  \bibinfo{author}{A.~Pal}, \bibinfo{author}{C.~Burgess},
  \bibinfo{author}{X.~Glorot}, \bibinfo{author}{M.~Botvinick},
  \bibinfo{author}{S.~Mohamed}, \bibinfo{author}{A.~Lerchner},
\newblock \bibinfo{title}{beta-vae: Learning basic visual concepts with a
  constrained variational framework}  (\bibinfo{year}{2016}).
\bibitem[{Mirza and Osindero(2014)}]{mirza2014conditional}
\bibinfo{author}{M.~Mirza}, \bibinfo{author}{S.~Osindero},
\newblock \bibinfo{title}{Conditional generative adversarial nets},
\newblock \bibinfo{journal}{arXiv preprint arXiv:1411.1784}
  (\bibinfo{year}{2014}).
\bibitem[{Gulrajani et~al.(2017)Gulrajani, Ahmed, Arjovsky, Dumoulin, and
  Courville}]{gulrajani2017improved}
\bibinfo{author}{I.~Gulrajani}, \bibinfo{author}{F.~Ahmed},
  \bibinfo{author}{M.~Arjovsky}, \bibinfo{author}{V.~Dumoulin},
  \bibinfo{author}{A.~Courville},
\newblock \bibinfo{title}{Improved training of wasserstein gans},
\newblock \bibinfo{journal}{arXiv preprint arXiv:1704.00028}
  (\bibinfo{year}{2017}).
\bibitem[{Lin et~al.(2020)Lin, Jain, Wang, Fanti, and Sekar}]{lin2020using}
\bibinfo{author}{Z.~Lin}, \bibinfo{author}{A.~Jain}, \bibinfo{author}{C.~Wang},
  \bibinfo{author}{G.~Fanti}, \bibinfo{author}{V.~Sekar},
\newblock \bibinfo{title}{Using gans for sharing networked time series data:
  Challenges, initial promise, and open questions},
\newblock in: \bibinfo{booktitle}{Proceedings of the ACM Internet Measurement
  Conference}, \bibinfo{year}{2020}, pp. \bibinfo{pages}{464--483}.
\bibitem[{Zhang et~al.(2020)Zhang, Yang, He, and Deng}]{zhang2020multimodal}
\bibinfo{author}{C.~Zhang}, \bibinfo{author}{Z.~Yang}, \bibinfo{author}{X.~He},
  \bibinfo{author}{L.~Deng},
\newblock \bibinfo{title}{Multimodal intelligence: Representation learning,
  information fusion, and applications},
\newblock \bibinfo{journal}{IEEE Journal of Selected Topics in Signal
  Processing} \bibinfo{volume}{14} (\bibinfo{year}{2020})
  \bibinfo{pages}{478--493}.
\bibitem[{Paletta et~al.(2021)Paletta, Arbod, and
  Lasenby}]{paletta2021benchmarking}
\bibinfo{author}{Q.~Paletta}, \bibinfo{author}{G.~Arbod},
  \bibinfo{author}{J.~Lasenby},
\newblock \bibinfo{title}{Benchmarking of deep learning irradiance forecasting
  models from sky images--an in-depth analysis},
\newblock \bibinfo{journal}{arXiv preprint arXiv:2102.00721}
  (\bibinfo{year}{2021}).
\bibitem[{Paletta and Lasenby(2020)}]{paletta2020convolutional}
\bibinfo{author}{Q.~Paletta}, \bibinfo{author}{J.~Lasenby},
\newblock \bibinfo{title}{Convolutional neural networks applied to sky images
  for short-term solar irradiance forecasting},
\newblock \bibinfo{journal}{arXiv preprint arXiv:2005.11246}
  (\bibinfo{year}{2020}).
\bibitem[{Stadt~Zurich(2021)}]{3Dstadt}
\bibinfo{author}{T.-u.~E. Stadt~Zurich}, \bibinfo{title}{3d city model},
  \bibinfo{year}{2021}. \URLprefix
  \url{https://www.stadt-zuerich.ch/ted/de/index/geoz/geodaten_u_plaene/3d_stadtmodell.html}.
\bibitem[{the OpenStreetMap~Foundation(2021)}]{OSM}
\bibinfo{author}{the OpenStreetMap~Foundation}, \bibinfo{title}{Openstreetmap},
  \bibinfo{year}{2021}. \URLprefix \url{https://www.openstreetmap.org/}.
\bibitem[{Ertugrul(2021)}]{@it}
\bibinfo{author}{E.~Ertugrul}, \bibinfo{title}{@it}, \bibinfo{year}{2021}.
  \URLprefix \url{https://www.food4rhino.com/en/app/it}.
\bibitem[{Kingma and Ba(2014)}]{kingma2014adam}
\bibinfo{author}{D.~P. Kingma}, \bibinfo{author}{J.~Ba},
\newblock \bibinfo{title}{Adam: A method for stochastic optimization},
\newblock \bibinfo{journal}{arXiv preprint arXiv:1412.6980}
  (\bibinfo{year}{2014}).
\bibitem[{Hinton et~al.(2012)Hinton, Srivastava, and
  Swersky}]{hinton2012neural}
\bibinfo{author}{G.~Hinton}, \bibinfo{author}{N.~Srivastava},
  \bibinfo{author}{K.~Swersky},
\newblock \bibinfo{title}{Neural networks for machine learning lecture 6a
  overview of mini-batch gradient descent},
\newblock \bibinfo{journal}{Cited on} \bibinfo{volume}{14}
  (\bibinfo{year}{2012}) \bibinfo{pages}{2}.
\bibitem[{Paszke et~al.(2019)Paszke, Gross, Massa, Lerer, Bradbury, Chanan,
  Killeen, Lin, Gimelshein, Antiga, Desmaison, Kopf, Yang, DeVito, Raison,
  Tejani, Chilamkurthy, Steiner, Fang, Bai, and Chintala}]{NEURIPS2019_9015}
\bibinfo{author}{A.~Paszke}, \bibinfo{author}{S.~Gross},
  \bibinfo{author}{F.~Massa}, \bibinfo{author}{A.~Lerer},
  \bibinfo{author}{J.~Bradbury}, \bibinfo{author}{G.~Chanan},
  \bibinfo{author}{T.~Killeen}, \bibinfo{author}{Z.~Lin},
  \bibinfo{author}{N.~Gimelshein}, \bibinfo{author}{L.~Antiga},
  \bibinfo{author}{A.~Desmaison}, \bibinfo{author}{A.~Kopf},
  \bibinfo{author}{E.~Yang}, \bibinfo{author}{Z.~DeVito},
  \bibinfo{author}{M.~Raison}, \bibinfo{author}{A.~Tejani},
  \bibinfo{author}{S.~Chilamkurthy}, \bibinfo{author}{B.~Steiner},
  \bibinfo{author}{L.~Fang}, \bibinfo{author}{J.~Bai},
  \bibinfo{author}{S.~Chintala},
\newblock \bibinfo{title}{Pytorch: An imperative style, high-performance deep
  learning library},
\newblock in: \bibinfo{editor}{H.~Wallach}, \bibinfo{editor}{H.~Larochelle},
  \bibinfo{editor}{A.~Beygelzimer}, \bibinfo{editor}{F.~d\textquotesingle
  Alch\'{e}-Buc}, \bibinfo{editor}{E.~Fox}, \bibinfo{editor}{R.~Garnett}
  (Eds.), \bibinfo{booktitle}{Advances in Neural Information Processing Systems
  32}, \bibinfo{publisher}{Curran Associates, Inc.}, \bibinfo{year}{2019}, pp.
  \bibinfo{pages}{8024--8035}. \URLprefix
  \url{http://papers.neurips.cc/paper/9015-pytorch-an-imperative-style-high-performance-deep-learning-library.pdf}.
\bibitem[{Abadi et~al.(2015)Abadi, Agarwal, Barham, Brevdo, Chen, Citro,
  Corrado, Davis, Dean, Devin, Ghemawat, Goodfellow, Harp, Irving, Isard, Jia,
  Jozefowicz, Kaiser, Kudlur, Levenberg, Man\'{e}, Monga, Moore, Murray, Olah,
  Schuster, Shlens, Steiner, Sutskever, Talwar, Tucker, Vanhoucke, Vasudevan,
  Vi\'{e}gas, Vinyals, Warden, Wattenberg, Wicke, Yu, and
  Zheng}]{tensorflow2015-whitepaper}
\bibinfo{author}{M.~Abadi}, \bibinfo{author}{A.~Agarwal},
  \bibinfo{author}{P.~Barham}, \bibinfo{author}{E.~Brevdo},
  \bibinfo{author}{Z.~Chen}, \bibinfo{author}{C.~Citro}, \bibinfo{author}{G.~S.
  Corrado}, \bibinfo{author}{A.~Davis}, \bibinfo{author}{J.~Dean},
  \bibinfo{author}{M.~Devin}, \bibinfo{author}{S.~Ghemawat},
  \bibinfo{author}{I.~Goodfellow}, \bibinfo{author}{A.~Harp},
  \bibinfo{author}{G.~Irving}, \bibinfo{author}{M.~Isard},
  \bibinfo{author}{Y.~Jia}, \bibinfo{author}{R.~Jozefowicz},
  \bibinfo{author}{L.~Kaiser}, \bibinfo{author}{M.~Kudlur},
  \bibinfo{author}{J.~Levenberg}, \bibinfo{author}{D.~Man\'{e}},
  \bibinfo{author}{R.~Monga}, \bibinfo{author}{S.~Moore},
  \bibinfo{author}{D.~Murray}, \bibinfo{author}{C.~Olah},
  \bibinfo{author}{M.~Schuster}, \bibinfo{author}{J.~Shlens},
  \bibinfo{author}{B.~Steiner}, \bibinfo{author}{I.~Sutskever},
  \bibinfo{author}{K.~Talwar}, \bibinfo{author}{P.~Tucker},
  \bibinfo{author}{V.~Vanhoucke}, \bibinfo{author}{V.~Vasudevan},
  \bibinfo{author}{F.~Vi\'{e}gas}, \bibinfo{author}{O.~Vinyals},
  \bibinfo{author}{P.~Warden}, \bibinfo{author}{M.~Wattenberg},
  \bibinfo{author}{M.~Wicke}, \bibinfo{author}{Y.~Yu},
  \bibinfo{author}{X.~Zheng}, \bibinfo{title}{{TensorFlow}: Large-scale machine
  learning on heterogeneous systems}, \bibinfo{year}{2015}. \URLprefix
  \url{https://www.tensorflow.org/}, \bibinfo{note}{software available from
  tensorflow.org}.
\bibitem[{Bartoli et~al.(1981)Bartoli, Coluzzi, Cuomo, Francesca, and
  Serio}]{bartoli1981autocorrelation}
\bibinfo{author}{B.~Bartoli}, \bibinfo{author}{B.~Coluzzi},
  \bibinfo{author}{V.~Cuomo}, \bibinfo{author}{M.~Francesca},
  \bibinfo{author}{C.~Serio},
\newblock \bibinfo{title}{Autocorrelation of daily global solar radiation},
\newblock \bibinfo{journal}{Il Nuovo Cimento C} \bibinfo{volume}{4}
  (\bibinfo{year}{1981}) \bibinfo{pages}{113--122}.
\bibitem[{Szcze{\'s}niak et~al.(2022)Szcze{\'s}niak, Ang, Letellier-Duchesne,
  and Reinhart}]{szczesniak2022method}
\bibinfo{author}{J.~T. Szcze{\'s}niak}, \bibinfo{author}{Y.~Q. Ang},
  \bibinfo{author}{S.~Letellier-Duchesne}, \bibinfo{author}{C.~F. Reinhart},
\newblock \bibinfo{title}{A method for using street view imagery to
  auto-extract window-to-wall ratios and its relevance for urban-level
  daylighting and energy simulations},
\newblock \bibinfo{journal}{Building and Environment} \bibinfo{volume}{207}
  (\bibinfo{year}{2022}) \bibinfo{pages}{108108}.

\end{thebibliography}

\end{document}